\documentclass[review]{elsarticle}

\usepackage{lineno,hyperref}
\modulolinenumbers[5]

\journal{Journal of Neurocomputing}

\usepackage{times}
\usepackage{epsfig}
\usepackage{graphicx}
\usepackage{amsmath}
\usepackage{amssymb}

\usepackage{algorithm}
\usepackage{algpseudocode}

\usepackage{graphicx}
\usepackage{caption}
\usepackage{enumitem}
\usepackage{xspace}
\usepackage{xparse}
\usepackage{amsmath}

\usepackage{multirow}

\DeclareDocumentCommand\newstep{o}{%
\item\IfNoValueTF{#1}{}{#1 \textendash\xspace}}

\newlist{steps}{enumerate}{1}
\setlist[steps]{label=\textit{Step \arabic*:},leftmargin=*}

\usepackage{pdfpages}
\usepackage{wrapfig}
\usepackage{array}
\newcolumntype{P}[1]{>{\centering\arraybackslash}p{#1}}

\usepackage{color,soul} 
\definecolor{tangelo}{rgb}{0.80, 0.2, 0.0}
\definecolor{black}{rgb}{0.0, 0.0, 0.0}

\newcommand{\R}{\mathbb{R}}

\newcommand{\comment}[1]{ }

\newcommand{\relmiddle}[1]{\mathrel{}\middle#1\mathrel{}}

\usepackage{grffile}
\usepackage{algorithm,algcompatible,lipsum}
\usepackage{mathtools}
\usepackage{comment}

\newcommand{\nr}{\mathcal{P}} 
\newcommand{\tr}{\mathcal{Q}} 

\begin{document}

\begin{frontmatter}

\title{Regularization in network optimization via \\ trimmed stochastic gradient descent with noisy label}

%
%
%
%

\author{Kensuke Nakamura$^\text{1}$}
\author{Bong-Soo Sohn$^\text{1}$}
\author{Kyoung-Jae Won$^\text{2}$}
\author{Byung-Woo Hong$^\text{1}$\corref{mycorrespondingauthor}}

\address{[1] Computer Science Department, Chung-Ang University, Seoul, Korea}
\address{[2] Biotech Research and Innovation Centre (BRIC), University of Copenhagen, Denmark}

\cortext[mycorrespondingauthor]{Corresponding author: Byung-Woo Hong}

\begin{abstract}
Regularization is essential for avoiding over-fitting to training data in network optimization, leading to better generalization of the trained networks. The label noise provides a strong implicit regularization by replacing the target ground truth labels of training examples by uniform random labels.  However, it can cause undesirable misleading gradients due to the large loss associated with incorrect labels.  We propose a first-order optimization method (Label-Noised Trim-SGD) that uses the label noise with the example trimming in order to remove the outliers based on the loss. The proposed algorithm is simple yet enables us to impose a large label-noise and obtain a better regularization effect than the original methods. The quantitative analysis is performed by comparing the behavior of the label noise, the example trimming, and the proposed algorithm.  We also present empirical results that demonstrate the effectiveness of our algorithm using the major benchmarks and the fundamental networks, where our method has successfully outperformed the state-of-the-art optimization methods.
\end{abstract}


\end{frontmatter}


%
%
%
%
%
%
\section{Introduction}
The neural networks learning is a large scale problem that is characterized by large size data-set with large model.
The neural network model consists of a number of layers that is known to approximate linear and non-linear functions.
Due to its high degree of freedom, however, the network network model is always at the risk of over-fitting to the training examples that degenerates the generalization, or the estimation performance for unknown data
Thus, regularization is required in the training process of the neural networks for better generalization.
\par
The neural network model is trained using the stochastic gradient descent (SGD) and its variants in combination with explicit and implicit regularization methods.
The explicit regularization restricts the model parameters with a prior knowledge,
e.g., weight-decay adds the regularization term into the object function, 
assuming the model parameters should follow a $L^2$-ball.
Dropout may assume that the model is an ensemble of sparse networks.
In contrast, the implicit methods offer regularization effect independent to the model structure.
SGD is actually an implicit regularization that updates the model using a subset of the training examples in an iterative manner that imposes the stochastic noise into the optimization process.
Early stopping is also an implicit regularization.
The label noise~\cite{edgington2007randomization,zhang2016understanding} is an implicit method that replaces the target label of randomly-selected examples by random uniform labels.
The label noise is simple and computationally efficient, yet it provides a strong regularization effect~\cite{zhang2016understanding} in the classification problems.
However, we find that the label noise also cause outliers with high loss values that can degenerate the model training.
\par
We propose a first-order optimization algorithm, called Label-Noised Trim-SGD, that intendedly uses the label noise with the example trimming in order to obtain an implicit regularization effect.
Our algorithm imposes the label noise and then removes data with low and high loss values using an example-trimming in order to remove outlier examples.  
This enables us to apply a large amount of label noise than the naive label-noise method, resulting in an improvement of generalization of the network model.
Different with the data trimming algorithms, we intentionally use the label noise in order to improve generalization of model.
\par
We relate our method to prior works in Section~\ref{sec:related_studies}
and present the naive algorithms of the label noise and the example trimming in Section~\ref{sec:preliminary}, followed by our proposed algorithm in Section~\ref{sec:ours}.  The effectiveness of our algorithm is demonstrated by experimental results in Section~\ref{sec:results} and
we conclude in Section~\ref{sec:conclusion}.
%
%
%
%
%
%
%
%
\section{Related works} \label{sec:related_studies}
\subsection{Explicit regularization}
%
\noindent {\bf Weight-decay:} 
is an explicit method that is equivalent to a $L^2$-norm regularization that injects the norm of model parameters in the objective function in order to penalize large parameter values.
The amount of weight decay is defined by the coefficient that is tuned by hand~\cite{simonyan2014very,chollet2017xception}, or learned by Bayesian optimization~\cite{snoek2015scalable,shahriari2016unbounded}.  The layer-wise and parameter-wise weight decay were also considered ~\cite{ishii2017layer,bengio2012practical,nakamura2019adaptive}.
The main drawback of the explicit regularization is that it depends on the architecture of model.
We in this paper focus on an implicit regularization
\par
\noindent {\bf Dropout:} is in particular used with classical shallow networks.
The dropout zeros the activation of randomly selected nodes with a certain probability during the training process~\cite{srivastava2014dropout}.
The dropping rate is generally set to be constant but its variants have been considered with adaptive rates depending on parameter value~\cite{ba2013adaptive}, estimated gradient variance~\cite{kingma2015variational}, biased gradient estimator~\cite{srinivas2016generalized}, layer depth~\cite{huang2016deep}, or marginal likelihood over noises~\cite{noh2017regularizing}.
However, in fact, the random masking to nodes can be erroneous to sparse models and the dropout is seldom used in recent studies.
%
%
%
%
%
%
\subsection{Implicit regularization}
%
\noindent {\bf SGD with annealing:}
The stochastic gradient~\cite{robbins1951stochastic,rumelhart1988learning,zhang2004solving,bottou2010large,bottou2018optimization}, calculated using a subset of data, gives a stochastic noise to gradient and provides an implicit regularization effect~\cite{zhu2019anisotropic}.
In SGD, parameters are updated by subtracting the gradient with the stochastic noise multiplied by the learning rate.
The learning rate should shrink in order to reduce the noise and converge the algorithm.
To this aim, a variety of learning rate annealing, e.g. exponential~\cite{george2006adaptive} and staircase~\cite{smith2017don}, and the adaptive learning rates, e.g., AdaGrad~\cite{duchi2011adaptive}, have been proposed, 
The sophisticated adaptive techniques, e.g., RMSprop~\cite{tieleman2012lecture} and Adam~\cite{kingma2014adam}, enable parameter-wise control of the learning rates.
The drawback of adaptive learning-rate techniques on the regularization is that they are practically inferior to SGD with scheduled annealing.
\par
\noindent {\bf Early stopping:}
The early stopping~\cite{prechelt1998early,zhang2016understanding} is a technique
of implicit regularization
where the training process is terminated manually at a certain epoch before the test loss increases.
In our experiments, we employ the learning-rate annealing using a sigmoid function that guarantees the convergence of 
the training loss within the specified epochs.
\par
\noindent {\bf Label noise:}
The label noise~\cite{edgington2007randomization,zhang2016understanding} is another implicit method that selects a subset of example randomly and replaces their labels by uniform random labels. 
Compared to the explicit methods, the label noise is simple to use yet it has a strong regularization effect~\cite{zhang2016understanding}.
However, as we observe in this study, the label noise can cause outliers with extreme loss that degenerate the model.
To alleviate this issue, we introduce the example-wise trimming based on the loss.
%
%
%
\subsection{Data trimming}
%
\noindent {\bf Byzantine robust optimization in distributed training:}
Distributed training of neural network models has become popular for accelerating the optimization process~\cite{dean2012large,goyal2017accurate}, where the model is updated using the mean of gradients computed by workers using given data. 
However bad worker or Byzantine attacks the mean gradient.  
Thus data-trimming SGD (trim-SGD) and other variants ~\cite{alistarh2018byzantine,yin2018byzantine,chen2019distributed} have been studied empirically and theoretically in order to remove the Byzantine failures.
\par
\noindent {\bf Example trimming for robust optimization:}
In the large-scale training, a bad example also degenerates the model optimization.
The example-wise and batch-wise trimming~\cite{feng2017outlier,shen2018learning,zhou2018hardThresholding,chi2019median} have been studied 
in order to remove such outliers.
Obviously these methods assume the existence of bad examples in the training data.
\par
Our algorithm is different with these studies in concept, i.e.,
we combine the label noise with example trimming in order to obtain a regularization effect 
that improves generalization of model.
\subsection{Optimization methods}
\vspace{3pt}
\noindent {\bf Energy landscape:}
The geometrical property of energy surface is helpful in optimization of highly complex non-convex problems associated with network architecture.
It is preferred to drive a solution toward local minima on a flat energy surface that is considered to yield better generalization~\cite{hochreiter1997flat,Chaudhari2017EntropySGD,dinh2017sharp} where flatness is defined around the minimum by its connected region, its curvature of the second order structure, and the width of its basin, respectively.
%
A geometry-driven variant of SGD has been developed in deep learning problems such as Entropy-SGD~\cite{Chaudhari2017EntropySGD}.
In our approach, we do not attempt to measure geometric property of loss landscape such as flatness with extra computational cost, but instead consider an implicit regularization.
\par
\vspace{3pt}
\noindent {\bf Variance reduction:} 
The variance of gradients is detrimental to SGD, motivating variance reduction techniques~\cite{Roux2012,johnson2013accelerating,Chatterji2018OnTT,zhong2014fast,shen2016adaptive,Difan2018svrHMM,Zhou2019ASim} that aim to reduce the variance incurred due to their stochastic process of estimation, and improve the convergence rate mainly for convex optimization while some are extended to non-convex problems~\cite{allen2016variance,huo2017asynchronous,liu2018zeroth}. 
One of the most practical algorithms for better convergence rates includes momentum~\cite{sutton1986two}, modified momentum for accelerated gradient~\cite{nesterov1983method}, and stochastic estimation of accelerated gradient (Accelerated-SGD)~\cite{Kidambi2018Acc}.
These algorithms are more focused on the efficiency in convergence than the generalization.
%
%
%
%
%

%
%
\section{Preliminary}  \label{sec:preliminary}
\subsection{Supervised learning with stochastic gradient descent}
Let us consider the general supervised classification problem first.
Given a set of examples each of which consists an input $x_i \in \R^m$ with the corresponding label $y_i \in \{1,...,L\} \quad (i=1,...,n)$, we train an estimation model $h_w(x_i)$ with the associated model parameters $w$ using a loss function 
%
\begin{eqnarray}
f_i(w) \coloneqq \l(h_w(x_i), y_i),
\end{eqnarray}
that measures the discrepancy between estimation $h_w(x_i)$ with the target label $y_i$.
The general supervised learning aims to find an optimal parameter $w^*$ that minimizes $\int_x \psi(x) \cdot \l(h_w(x_i), y_i) \,dx$ where $\psi(x)$ is the probability of $x$ in the true distribution.
Since $\psi(x)$ is unknown, we often use the average loss of examples
\begin{align}
    F(w) = \frac{1}{n} \sum_{i=1}^n \l(h_w(x_i), y_i) = \frac{1}{n} \sum_{i=1}^n f_i(w). \label{eq:energy}
\end{align}
When the loss function is differentiable and convex on parameter $w$, we can minimize Eq.(\ref{eq:energy}) using the gradient descent method that follows 
$w^{t+1} = w^{t} - \eta^t \cdot \nabla F(w^{t})$,
%
where $\nabla F(w^{t})$ is the loss gradient and $\eta^t$ is the learning-rate.
However this is computationally intractable in the large-scale problem where the number of examples is, e.g., tens of thousands.
Moreover $F(w)$ is usually non-convex in machine learning problems.  Thus the full gradient $\nabla F(w^{t})$ leads the model into a local minima and degenerates the generalization of the trained model or the applicability to unseen data.
\par
The common choice to alleviate these issues is the stochastic gradient descent (SGD) that updates model using a subset of examples ($\beta$), called mini-batch, as
\begin{align}
     w^{t+1} = w^{t} - \eta^t \left( \frac{1}{B} \sum_{i \in \beta^t} \nabla f_i(w^{t}) \right). \label{eq:sgd}
\end{align}
%
The batch size $B \coloneqq |\beta|$ is usually small and the complexity per iteration is free from the total number of data.  
Also SGD naturally introduces a stochastic noise that is known to help the model escape from local-and saddle-points~\cite{ge2015escaping,jin2017escape,Kleinberg2018AnAV}. 
\subsection{Label noise}
The label noise~\cite{edgington2007randomization,zhang2016understanding} is an implicit regularization technique 
that selects a subset of the training examples with probability $\nr$ 
and re-assigns their label by a uniform random class at each iteration as
\begin{align} \label{eq:label_noise}
    \hat{y}_i \coloneqq G(y_i; \nr) =
        \begin{cases}
        y_i   & \text{if $\chi_i(\nr) = 1$}, \\
        \Theta(L)  & \text{if $\chi_i(\nr) = 0$},
    \end{cases}
\end{align}
where $\hat{y}_i$ is the noised label, $\Theta (L) \in \{1, ..., L\}$ is random uniform within the label set, and $\chi(\nr)$ is a Bernoulli random variable, following $\Pr(\chi_i=1)=\nr$.
The baseline SGD is generalized by $\nr=0$.
\subsection{Example trimming}
There are a variety of data trimming techniques relating to gradient descent algorithm, e.g.,~\cite{dean2012large,goyal2017accurate,alistarh2018byzantine,yin2018byzantine,chen2019distributed,feng2017outlier,shen2018learning,zhou2018hardThresholding,chi2019median}.  
We use an example trimming in combination with SGD that we call Trim-SGD, selecting a subset of the training examples in mini-batch as
\begin{align} \label{eq:trimming}
\hat{\beta} \coloneqq  H(\beta; \tr) = \left \{ j \relmiddle| \frac{\tr}{2}\cdot B < \mathrm{order} (j) < (1 - \frac{\tr}{2}) \cdot B  \right \},
\end{align}
%
where $\mathrm{order} (j) \in \{1,...,B\}$ is the rank of example $j$ in the loss at that iteration,
i.e., we prune the top-$\tr/2$ and bottom-$\tr/2$ examples using the loss and obtain $\hat{B} = (1 - \tr) \cdot B$ examples, where $\tr=0$ generalizes the vanilla SGD.
%
%
%
%
%
%
%
%
%
%
%
%
%
\section{Proposed algorithm}  \label{sec:ours}
\subsection{Distribution of loss with noised labels}
\begin{figure} [htb]
\centering
\includegraphics[width=0.65\textwidth]{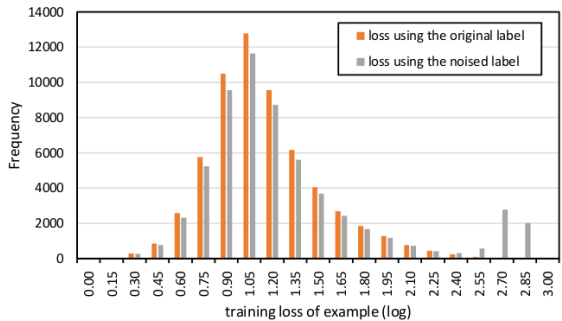}\\
\vspace{-5pt}
\caption{The frequency of example loss (x-axis in log scale) using the original label (orange) and the noised label (gray) with $\nr=10\%$ for MNIST by NN-2 trained for 50 epochs using the original labels.}
\label{fig:loss_distribution_by_noise}
\end{figure}

The presented algorithm is motivated by our observation on the distribution of loss with the label noise.
Figure~\ref{fig:loss_distribution_by_noise} presents 
(y-axis) the number of examples over (x-axis) their loss computed using the original label and the noised label given by Eq.(\ref{eq:label_noise}) with $\nr=10\%$,
where we trained a neural network~\cite{blum1991approximation} for MNIST~\cite{lecun1998gradient} using SGD for 50 epochs under a basic condition where a fixed learning-rate of $\eta=0.01$ without momentum and the batch size of $B=128$ were employed.  Then we calculated the loss of the 60K training examples using the original labels (orange), and those using the noised labels (gray).
\par
As demonstrated in Figure~\ref{fig:loss_distribution_by_noise}, there are outliers with extremely high loss due to the label noise while the label noise increases and also decreases the loss of example.
The unusually high loss is harmful in the model update since we use the mean loss of examples.
Our assumption is that if we remove these outliers, we can improve that generalization of the trained model.
To this aim, we employ the example trimming that prunes examples using loss.  
%
%
%
%
%
%
%
\subsection{Trimmed SGD with noisy labels}
\par
We propose a first-order optimization algorithm, namely Label-Noised Trim-SGD, that combines the label noise with the example trimming.
Our algorithm aims to obtain a mild regularization effect 
by removing outlier examples with low and high loss while imposing the label noise to data. 
This guarantees that the trimmed loss is oriented to the median of the example distribution irrespective of the choice of the trimming ratio.
Formally, our training loss is given as
\begin{align} \label{eq:our_loss}
\hat{f}_i(w) \coloneqq \l(h_w(x_i), \hat{y}_i),
\end{align}
where $\hat{y}_i$ is the noised label given by Eq.(\ref{eq:label_noise}), and the model is updated by
\begin{align}
     w^{t+1} = w^{t} - \eta^t \left( \frac{1}{\hat{B}} \sum_{i \in \hat{\beta}^t} \nabla \hat{f}_i(w^{t}) \right), \label{eq:our_update}
\end{align}
where again $\hat{\beta}$ is a subset of mini-batch examples trimmed by Eq.(\ref{eq:trimming}) and $\hat{B}$ is the number of examples in $\hat{\beta}$.
\par
Algorithm~\ref{alg:ours} presents the pseudo code of our algorithm.
Given a mini-batch, we apply the label noise to the examples and compute the loss of each training example with the noised labels.
We then trim examples using their loss before computing gradients by back-propagating the loss of the trimmed examples.
The example-wise loss can be computed in parallel, and also the back-propagation and the model update of our method are the same with the original SGD.  Thus the additional cost that our algorithm requires is incurred for sorting the examples that has the complexity of $O\left(B\log(B)\right)$, where $B$ is the batch size.  Since $B$ is small in usual, the additional cost is quite small compared with the cost of the back-propagation. 
%
%
%
%
%
%
%
\begin{algorithm}[htb]
\caption{\small Label-Noised Trim-SGD}
\label{alg:ours}
\begin{algorithmic}
\small
\STATE $\{x_j, y_j\}^t_B$: a mini-batch of input $x_j$ and its target $y_j$ at iteration $t$, $j = \{1, ..., B\}$
\STATE $\nr$ : hyper-parameter for the noise injection to targets
\STATE $\tr$ : hyper-parameter for the example trimming
\vspace{8pt}
	\FORALL {$t$ : iteration} 
		\STATE inject a random noise to label using Eq.(\ref{eq:label_noise}) with $\nr$
		\STATE compute example loss with the noised labels
		\STATE trim the examples using Eq.(\ref{eq:trimming}) with $\tr$
		\STATE back-prop the loss gradient and update model as Eq.(\ref{eq:our_update}) 
	\ENDFOR
\vspace{3pt}
\end{algorithmic}
\end{algorithm}
%
%

%
%
\section{Experimental results} \label{sec:results}
%
We present empirical results on the the label noise and the example trimming, followed by the proposed algorithm.
We then provide quantitative evaluation of the proposed algorithm in comparison to the state-of-the-arts.
%
%
\par
In the experiments,we use three data-sets: MNIST~\cite{lecun1998gradient}, Fashion-MNIST~\cite{xiao2017online}, and and EMNIST-Letters~\cite{cohen2017emnist} that are preferred for testing algorithms in machine learning.  
MNIST is a conventional benchmark data-set that consists of $60$K training and $10$K test gray images of handwritten 10-digits.
Fashion-MNIST is a more difficult image classification data-set that consists of $60$K training and $10$K test images with 10 fashion item categories.
EMNIST-Letters (or EMNIST) is also a challenging benchmark that consists of about $125$K training and $5$K test images with the 26 letters of the Alphabet.
We employ these data-sets since they consist of gray images such that we can apply the same network model for the different types of image classification tasks.
%
%
%
%
%
We use the cross entropy loss as the typical loss function for the image classification task.
%
%
%
We use three of the fundamental network models: fully-connected neural networks with two hidden layers (NN-2) and with three hidden layers (NN-3)~\cite{blum1991approximation}, LeNet~\cite{lecun1998gradient} with two convolution layers followed by two of fully-connected layers.
%
\par
In our comparative analysis, we consider the following optimization algorithms: 
SGD, RMSprop~\cite{tieleman2012lecture} and Adam~\cite{kingma2014adam}, Entropy-SGD (eSGD)~\cite{Chaudhari2017EntropySGD}, Accelerated-SGD (aSGD)~\cite{Kidambi2018Acc}, and our algorithm (Ours).
%
%
We use SGD as the baseline for comparing the performance of the above algorithms of which the common hyper-parameters such as learning rate, batch size, and momentum are chosen with respect to the best test loss of the baseline.  
Specifically, we employ the sigmoid annealing in which the learning rate starts at $\eta$ and ends with $\eta / 100$ for SGD, eSGD, aSGD, and ours.  RMSprop and Adam determine the step size dynamically with the initial learning rate $\eta$.
The initial learning rate is chosen within $\{$0.005, 0.01, 0.05, 0.1$\}$ for SGD and ours, 
$\{$0.0001, 0.0005, 0.001, 0.005$\}$ for RMSprop and Adam, $\{$0.01, 0.05, 0.1,0.5$\}$ for eSGD, and $\{$0.001, 0.005, 0.01, 0.05$\}$ for aSGD, depending on data-set and network model in order to make experimental results independent to the scale of initial learning rate.
We use the batch size of $B=128$, the momentum of 0.9, and the epoch size of 100 as a practical condition.
%
Most of our experiments are performed without the weight-decay since the role of explicit regularizer is to support the insufficient regularization of network architecture and is different with implicit ones~\cite{zhang2016understanding}.
%
%
\par
In all of the experiments, we perform 10 independent trials and their average learning curves 
indicating the training-and test-loss are considered. %
In particular, we consider the mean of test loss in the last $10\%$ epochs (mean test loss) as well as
the minimum test loss across all the epochs and the trials (minimum test loss).
The mean loss is the objective in our problem of Eq.(\ref{eq:energy}) and thus is the general performance measure of algorithms.  The minimum loss presents the convergence point with respect to the test loss curve and is useful for evaluating the potential performance of algorithms.
Note that both the label noise and  the trimming are applied only to the training examples. 
%
In following figures and tables, we scale the loss value for MNIST by 100, and Fashion-MNIST and EMNIST by 10  for the numerical convenience. 
%
%
%

%
%
%
%
\subsection{Preliminary experiments}
\vspace{3pt}
\noindent {\bf Effect of label noise:} 
We revisit the regularization by label noise~\cite{edgington2007randomization,zhang2016understanding}.
Figure~\ref{fig:pretest:noise} presents the training and test loss curves for MNIST by NN-2 and LeNet trained using SGD with the label noise of $\tr=0$ and $\tr=2.5\%$ under (top) the basic condition in which the learning-rate is fixed at $\eta=0.01$, the momentum and weight-decay are not applied, and the epoch size is extended to 500 epochs;
(middle) the same condition but the weight decay is applied, and (bottom) our practical condition where we use the sigmoid learning-rate annealing with the momentum.
As demonstrated in Figure~\ref{fig:pretest:noise} (top), the label noise increases the training loss (red dotted line) but makes it converge more quickly than the baseline SGD with $\nr=0$.  
Meanwhile the label noise slows the convergence of the test loss (blue solid-line) or alleviates the over-fitting of model, but increases the test loss at the convergence point.
(middle) The label noise is still effective when combined with the weight decay, since the implicit regularization is different with the explicit ones~\cite{zhang2016understanding}.
(bottom) The label noise also shows the regularization effect under the practical condition.
%
%
%
%
%
\def\fw{80pt}
\def\pw{70pt}
\begin{figure} [htb]
\centering
\scriptsize
\includegraphics[width=120pt]{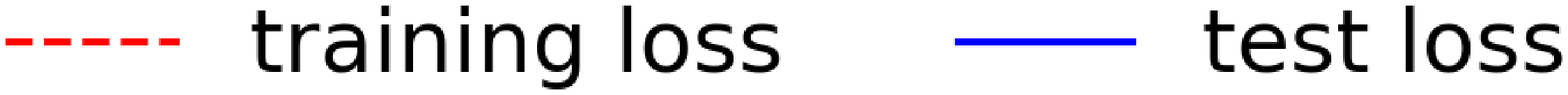} \\
\begin{tabular}{P{\pw}P{\pw}P{5pt}P{\pw}P{\pw}}
\includegraphics[width=\fw]{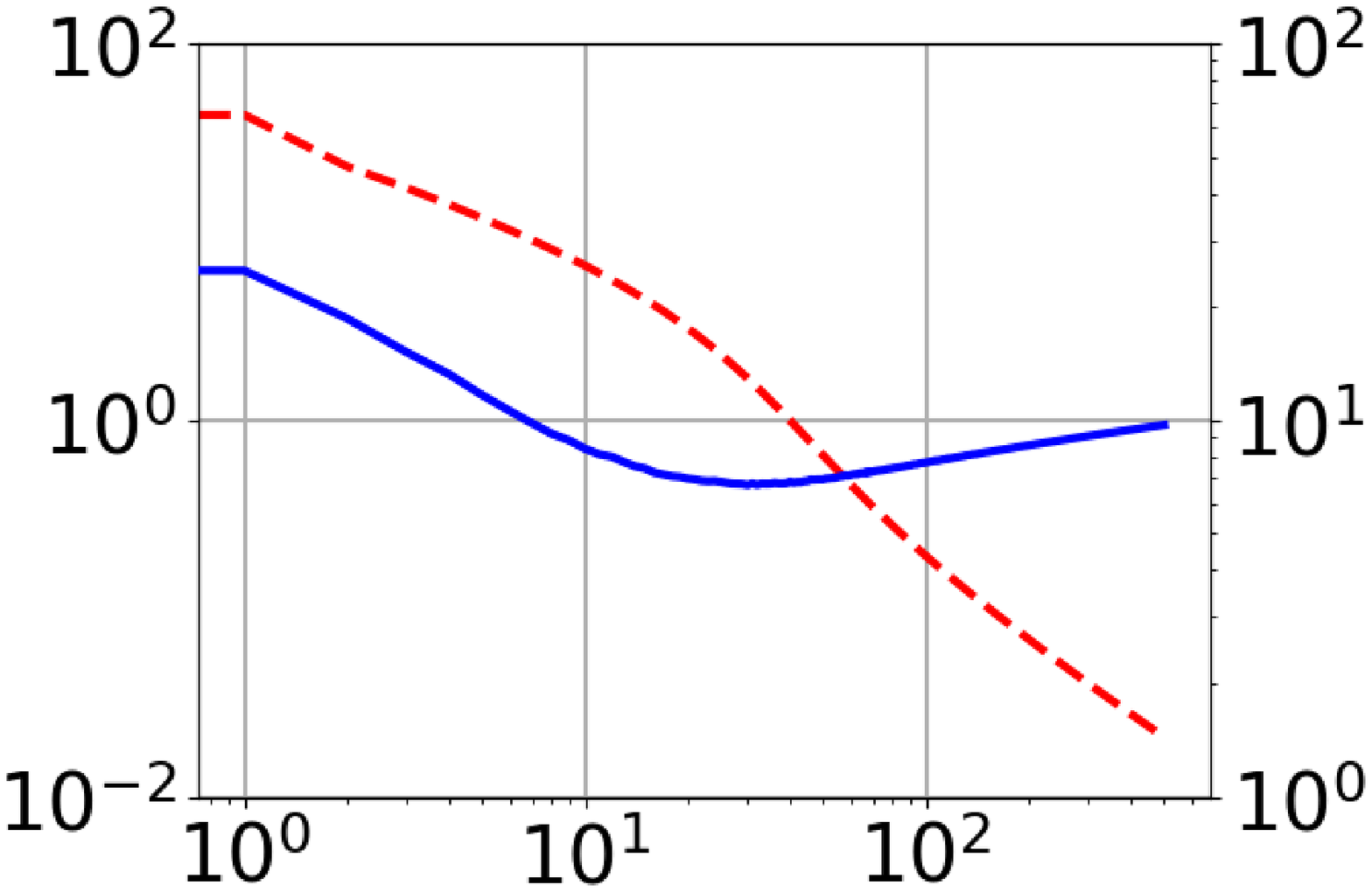} &
\includegraphics[width=\fw]{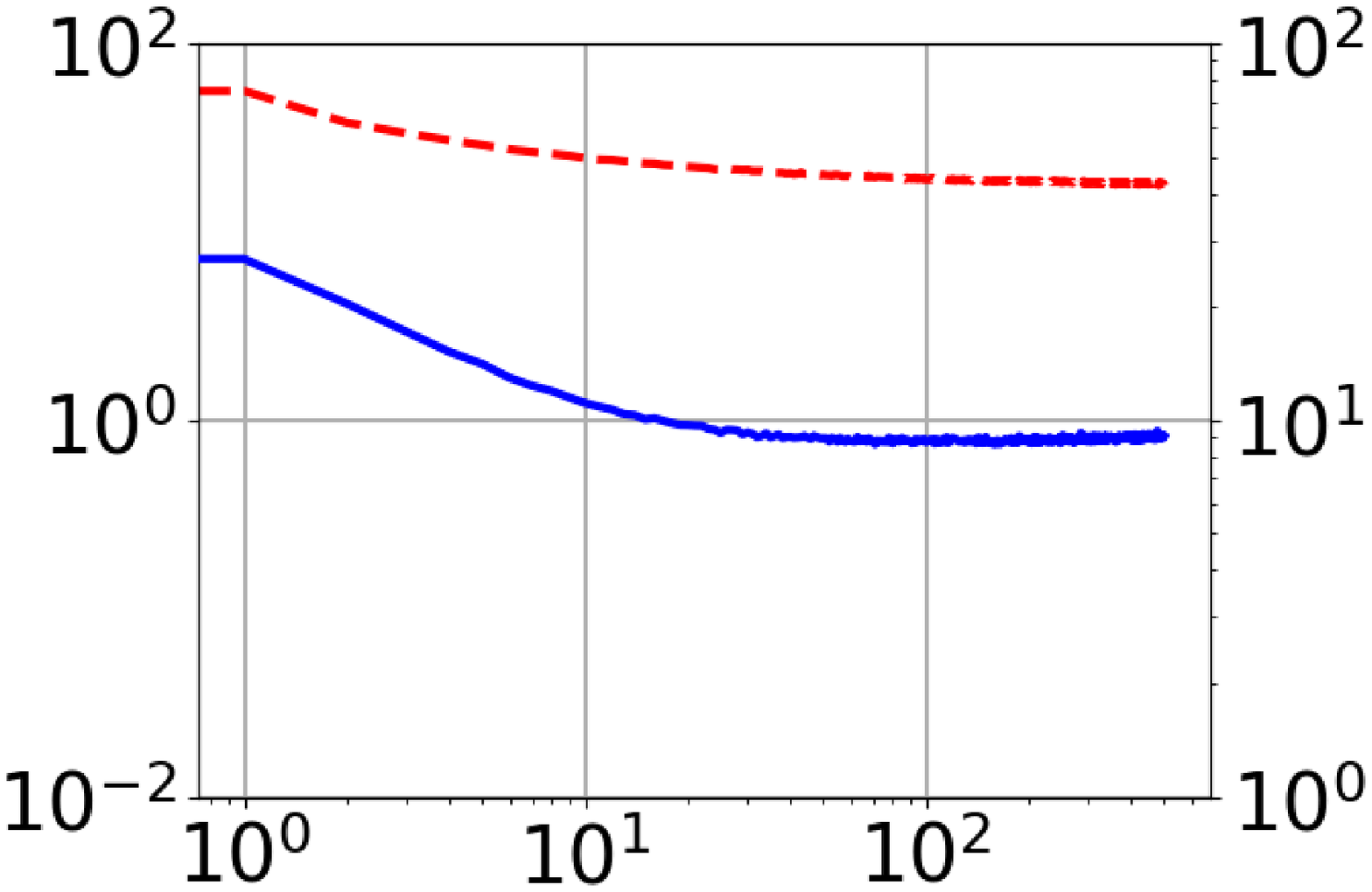} &
&
\includegraphics[width=\fw]{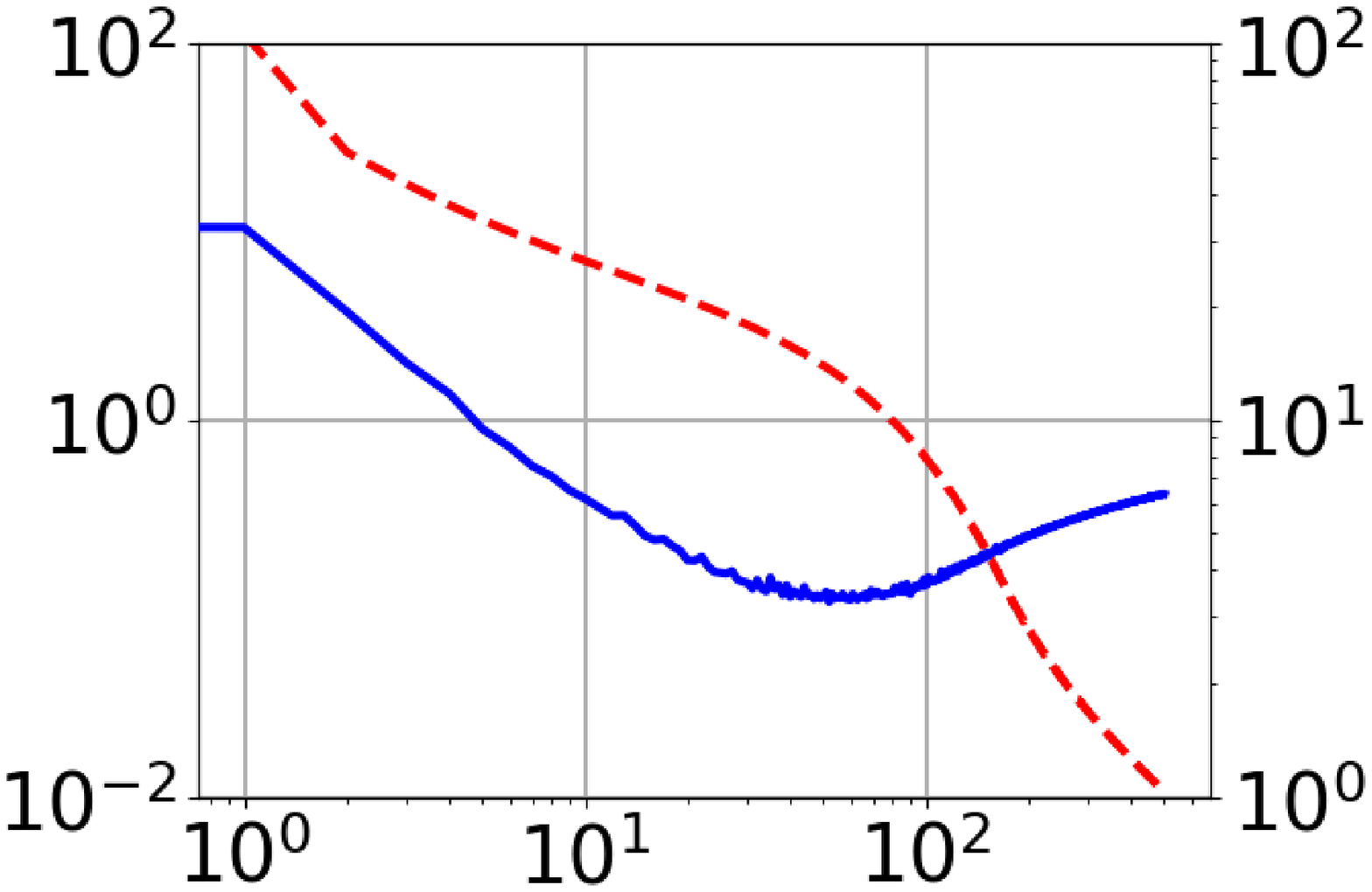} &
\includegraphics[width=\fw]{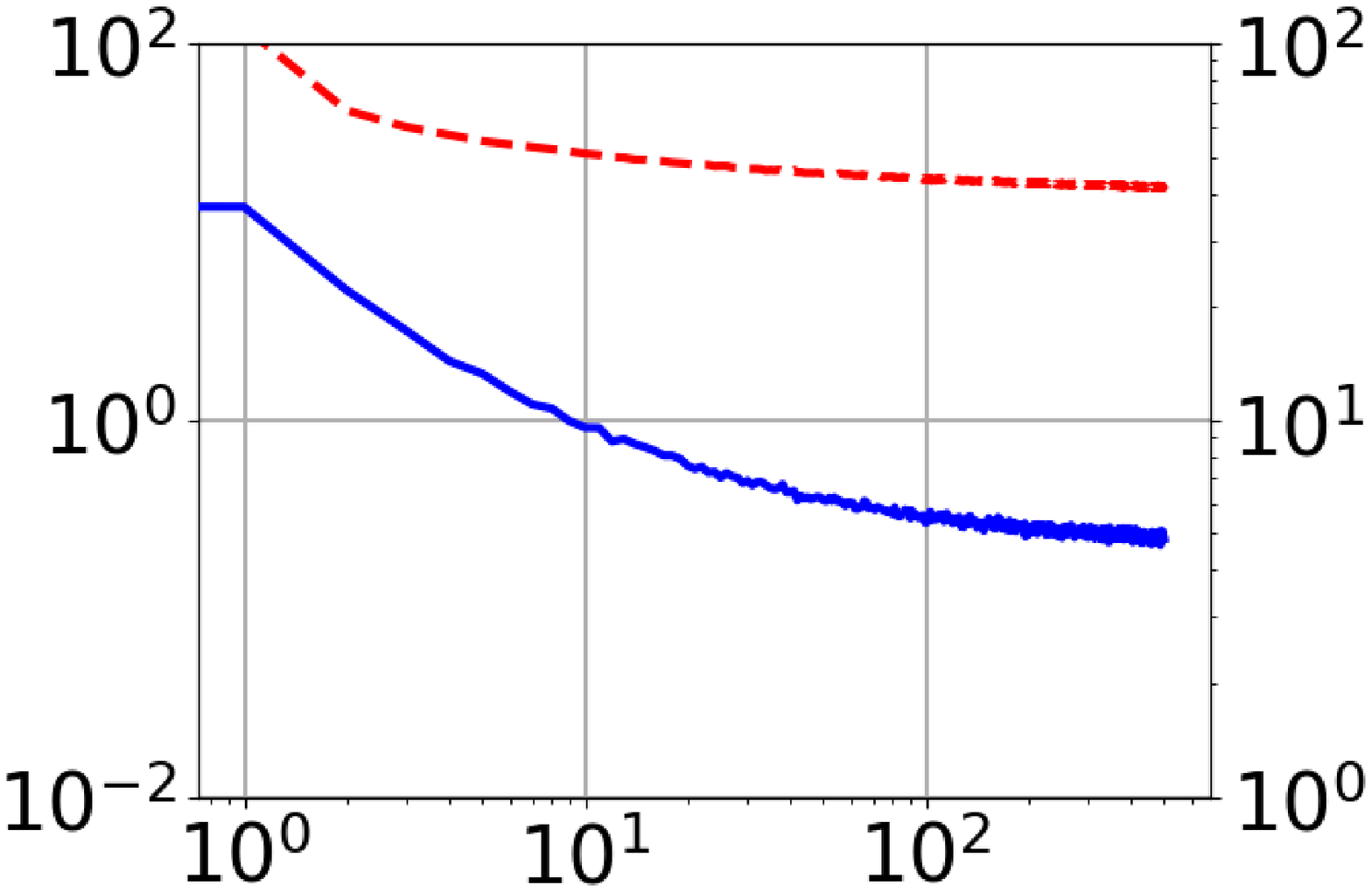} \\
\includegraphics[width=\fw]{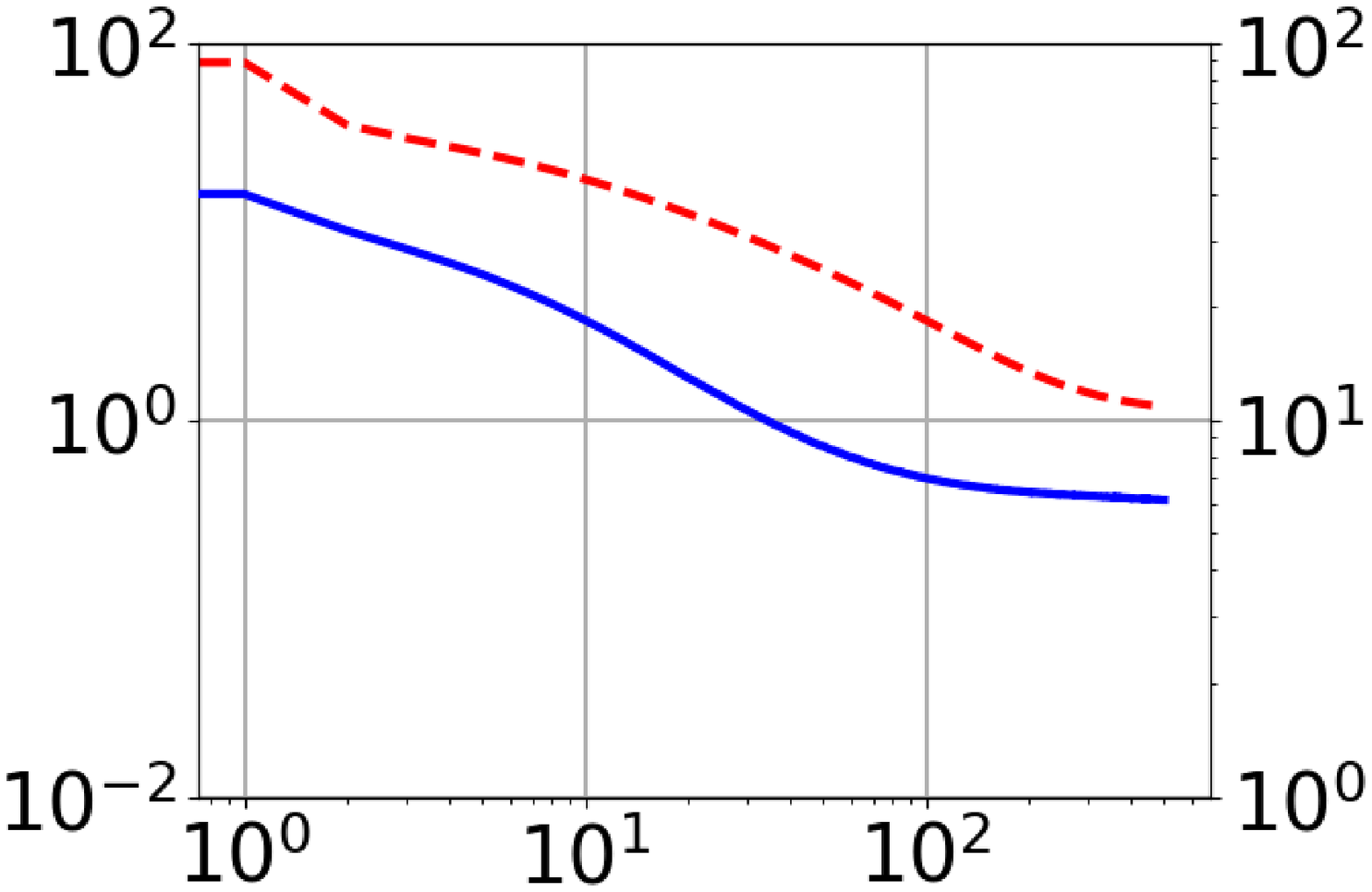} &
\includegraphics[width=\fw]{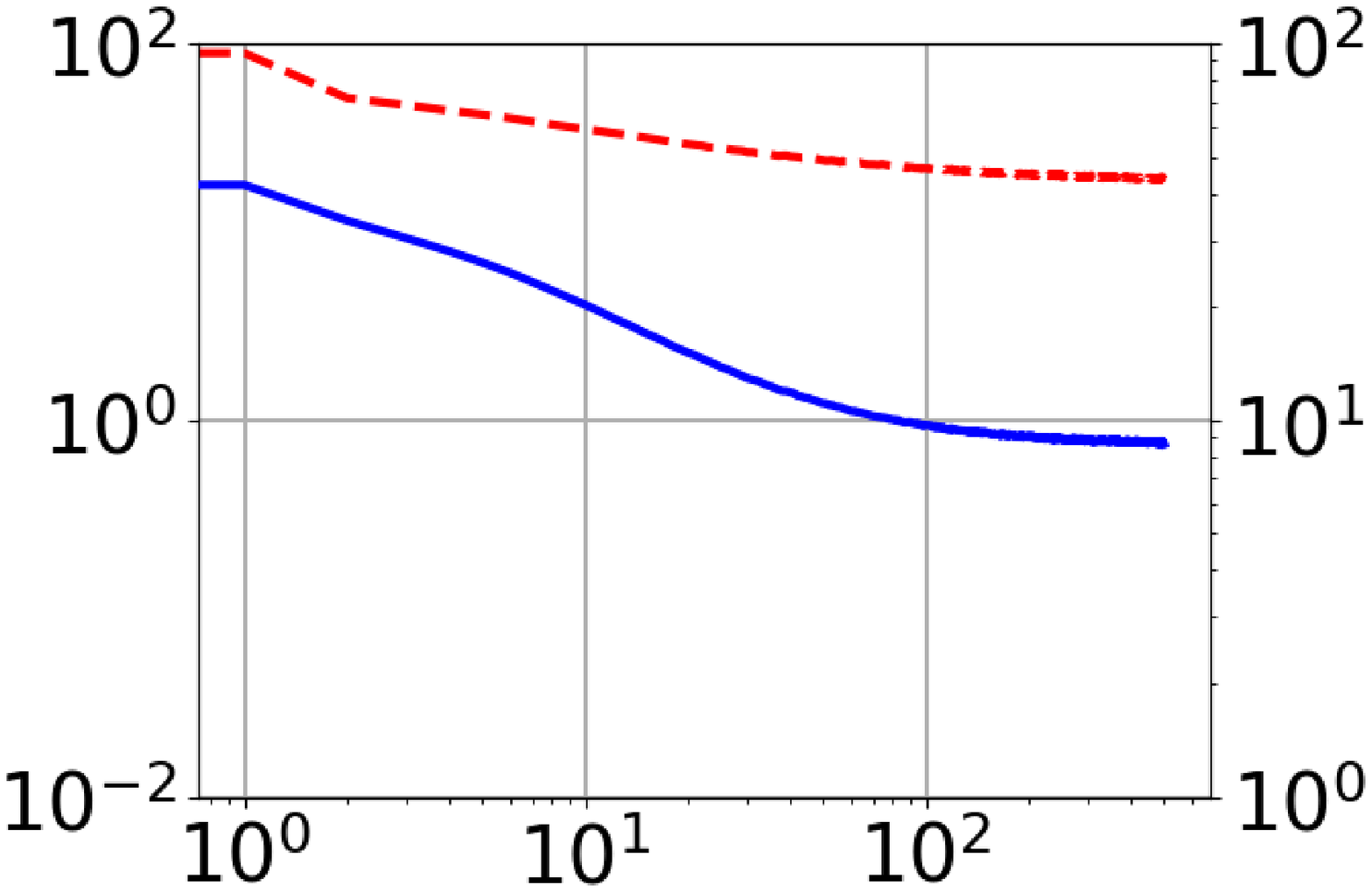} &
&
\includegraphics[width=\fw]{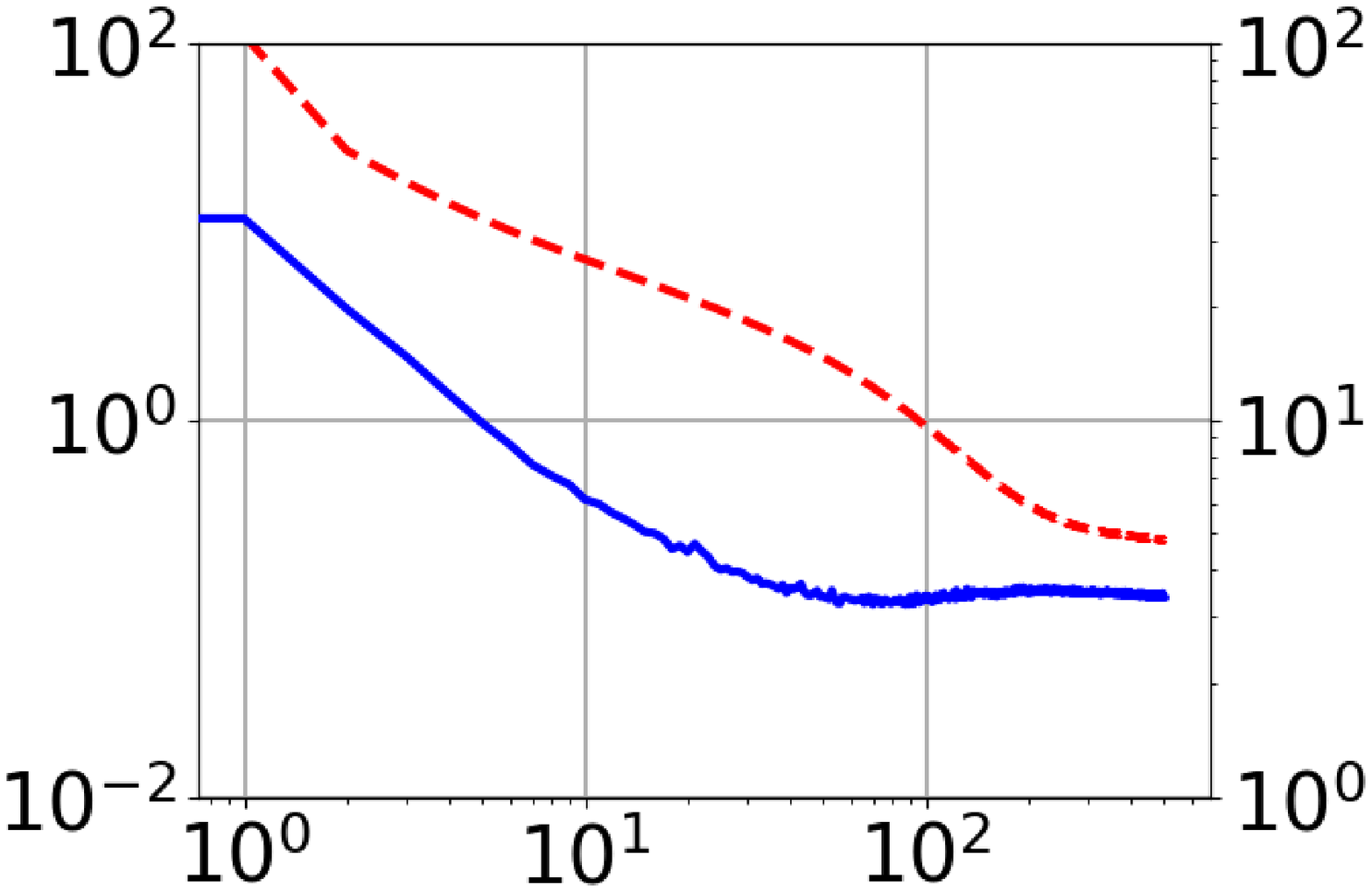} &
\includegraphics[width=\fw]{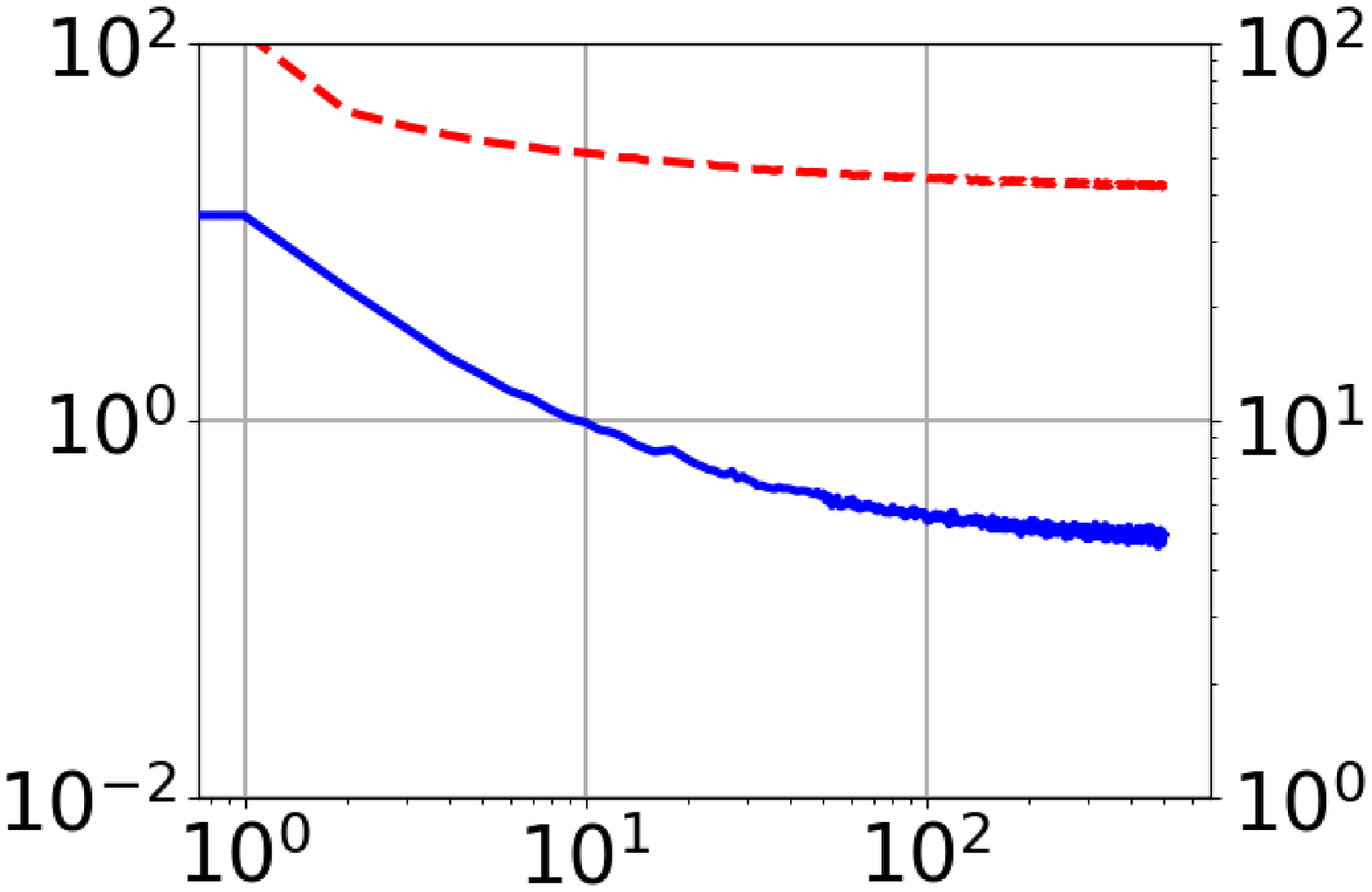} \\
\includegraphics[width=\fw]{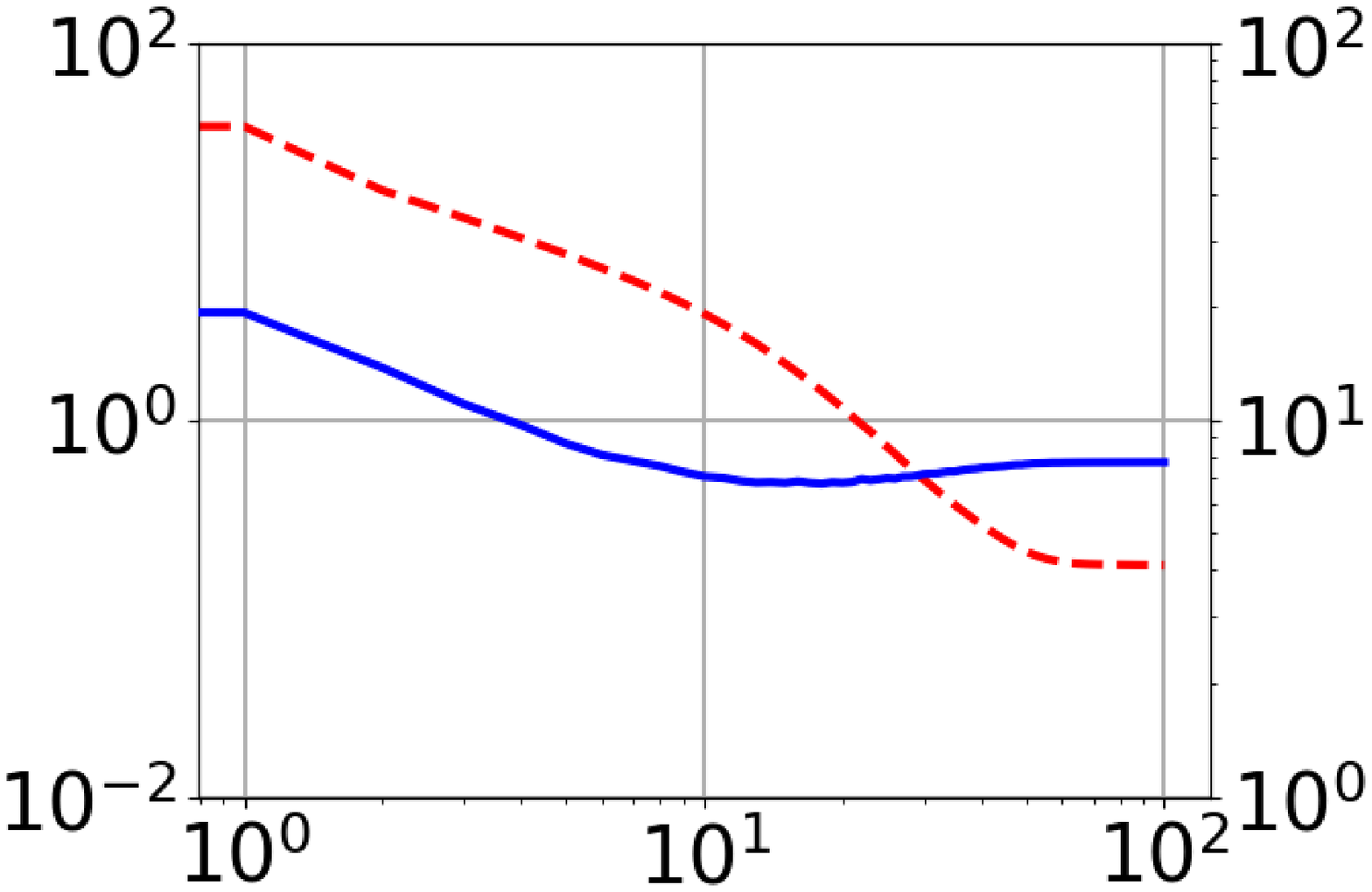} &
\includegraphics[width=\fw]{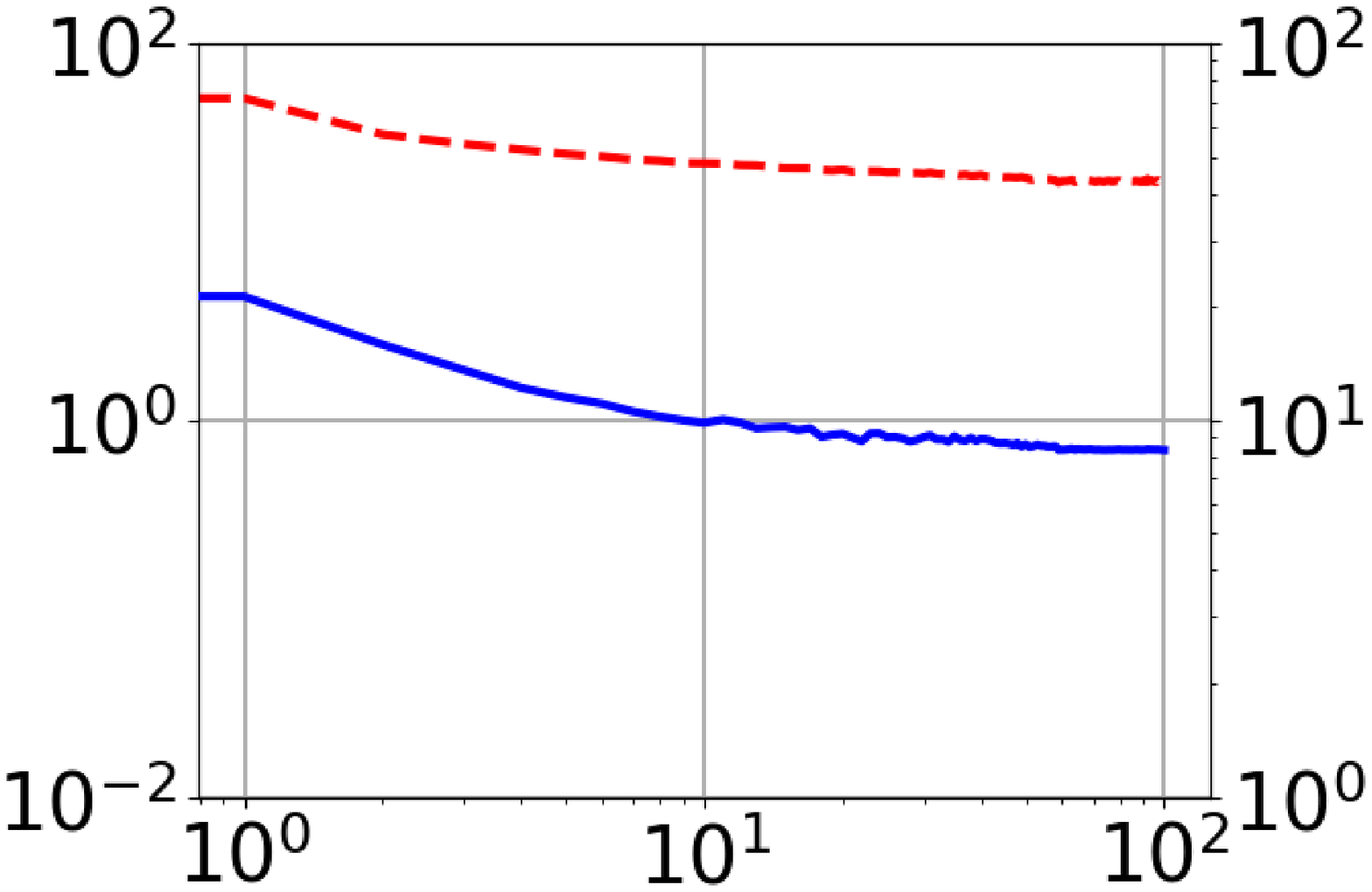} &
&
\includegraphics[width=\fw]{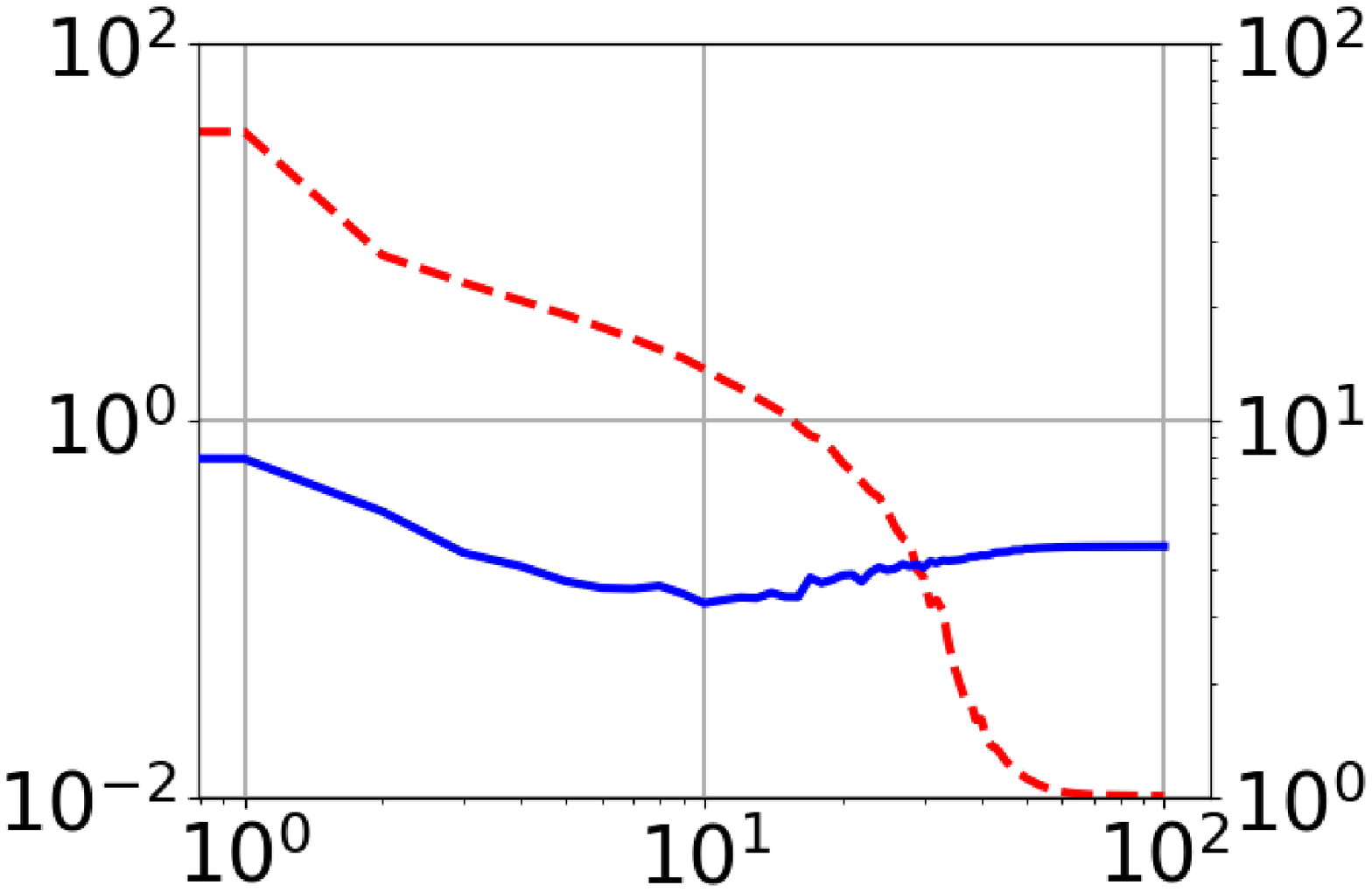} &
\includegraphics[width=\fw]{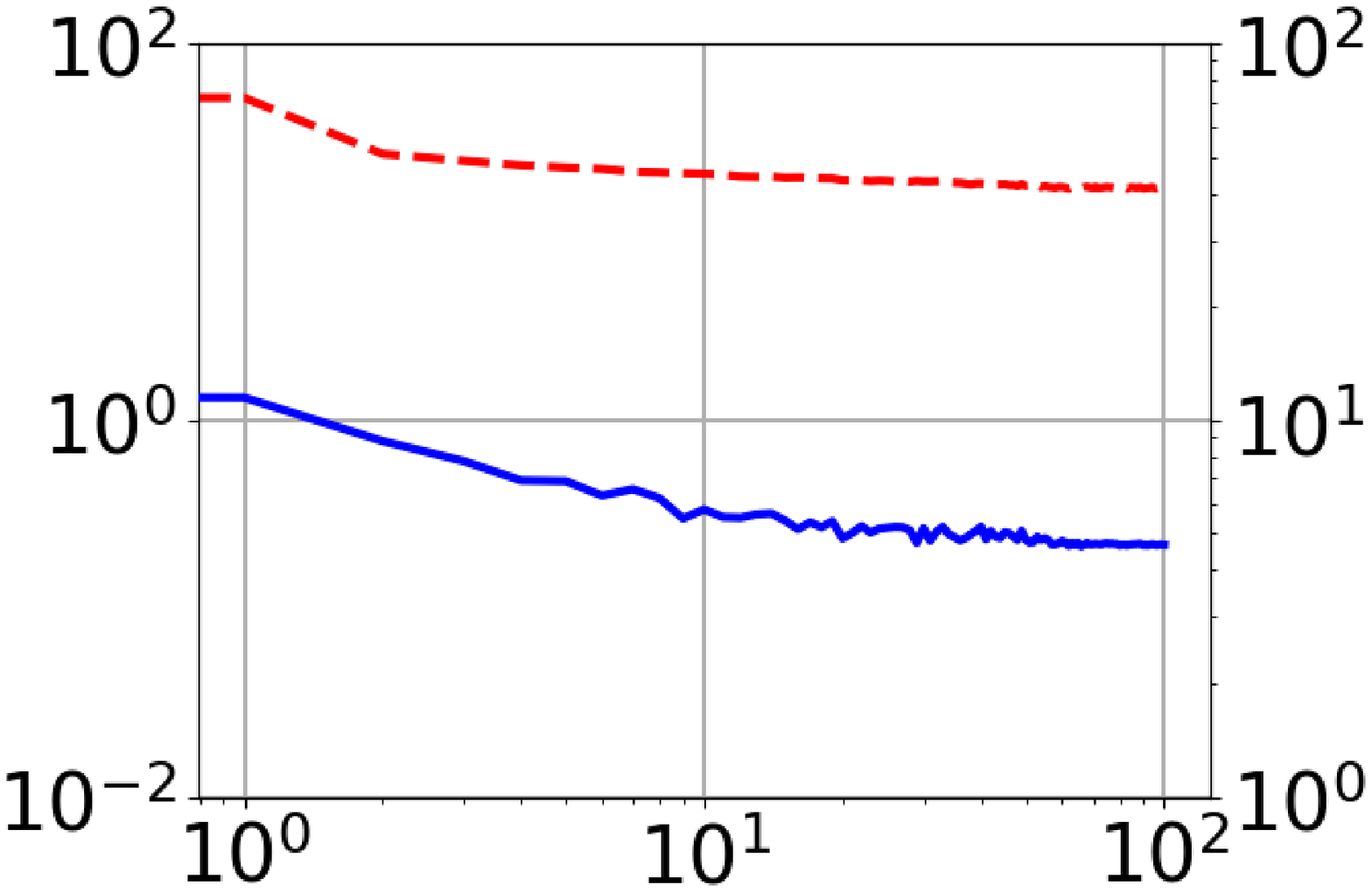} \\
\quad $\nr=0$ (baseline) &
\quad $\nr=2.5\%$ & 
&
\quad $\nr=0$ & 
\quad $\nr=2.5\%$ \\
\multicolumn{2}{c}{\small \quad (1) NN-2} && \multicolumn{2}{c}{\small \quad (2) LeNet}\\
\end{tabular}
\vspace{-5pt}
\caption{[Effect of label noise for SGD] 
Training loss (red dotted-line indicated by 1st y-axis) and test loss (blue solid-line indicated by 2nd y-axis) for MNIST over epoch (x-axis) by NN-2 (left) and LeNet (right) networks
trained using 
(top) a fixed learning-rate without momentum where the 
epoch size is extended to 500, (middle) the same basic condition with weight decay of $\lambda=5\times10^{-4}$,
and (bottom) the practical condition using sigmoid learning-rate with momentum.}
\label{fig:pretest:noise}
\end{figure}
%
%
%
%
%
%
%
%
%
%
\def\fw{80pt}
\def\pw{70pt}
\begin{figure} [htb]
\centering
\scriptsize
\includegraphics[width=180pt]{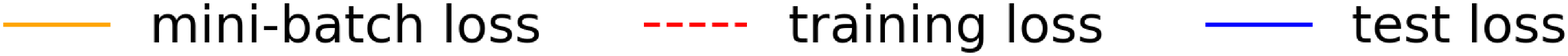} \\
\begin{tabular}{P{\pw}P{\pw}P{5pt}P{\pw}P{\pw}}
\includegraphics[width=\fw]{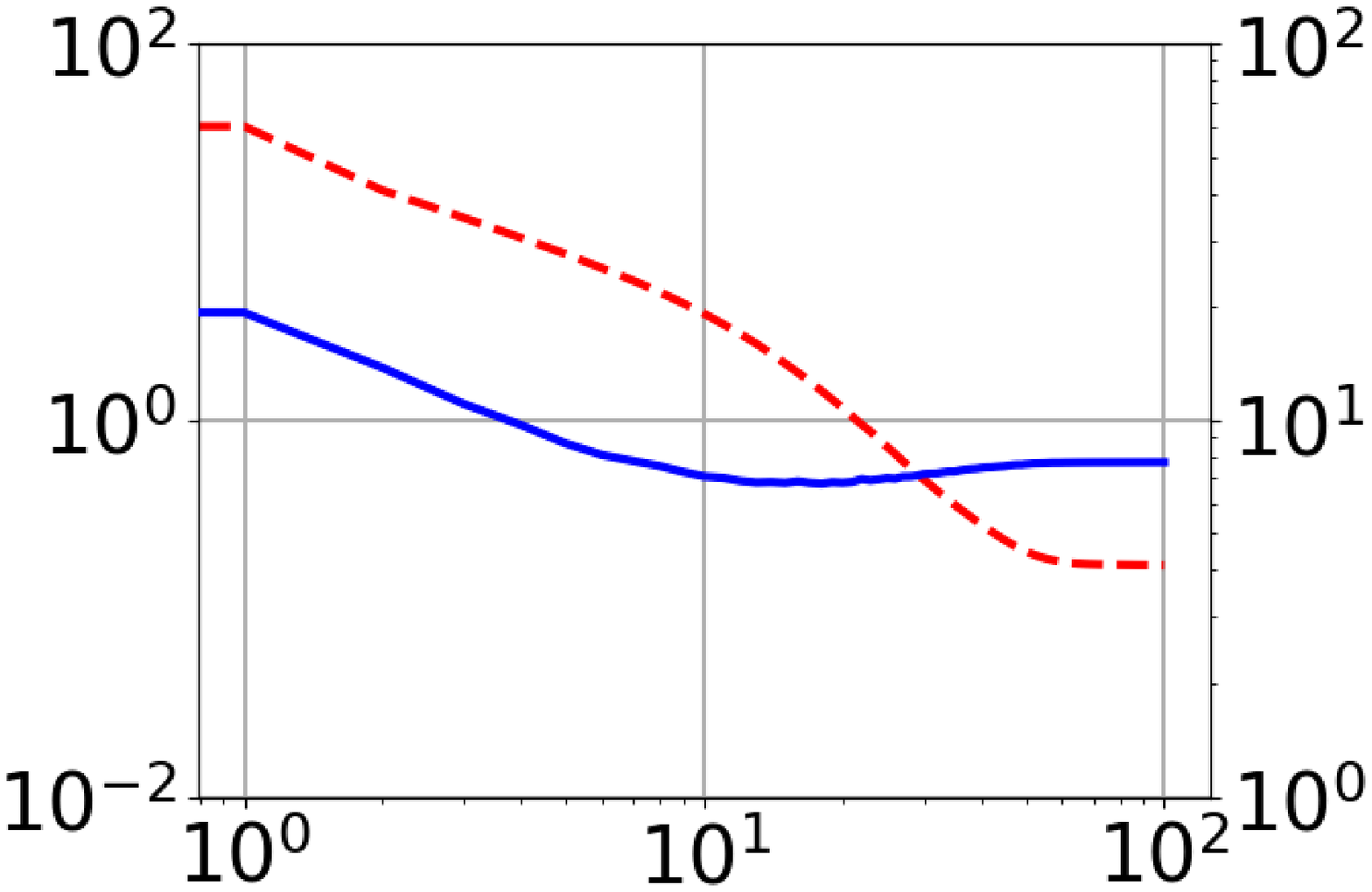} &
\includegraphics[width=\fw]{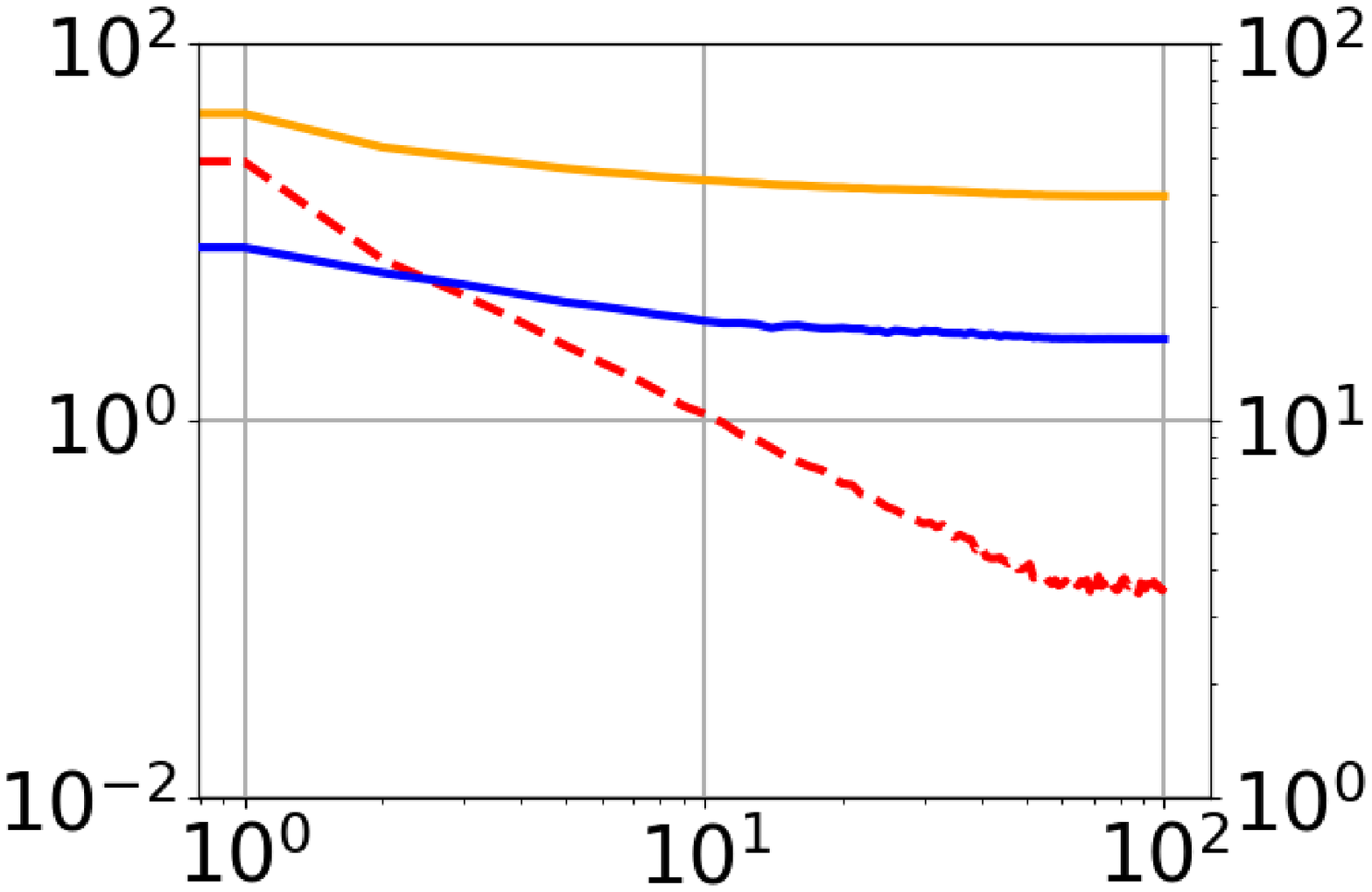} &
&
\includegraphics[width=\fw]{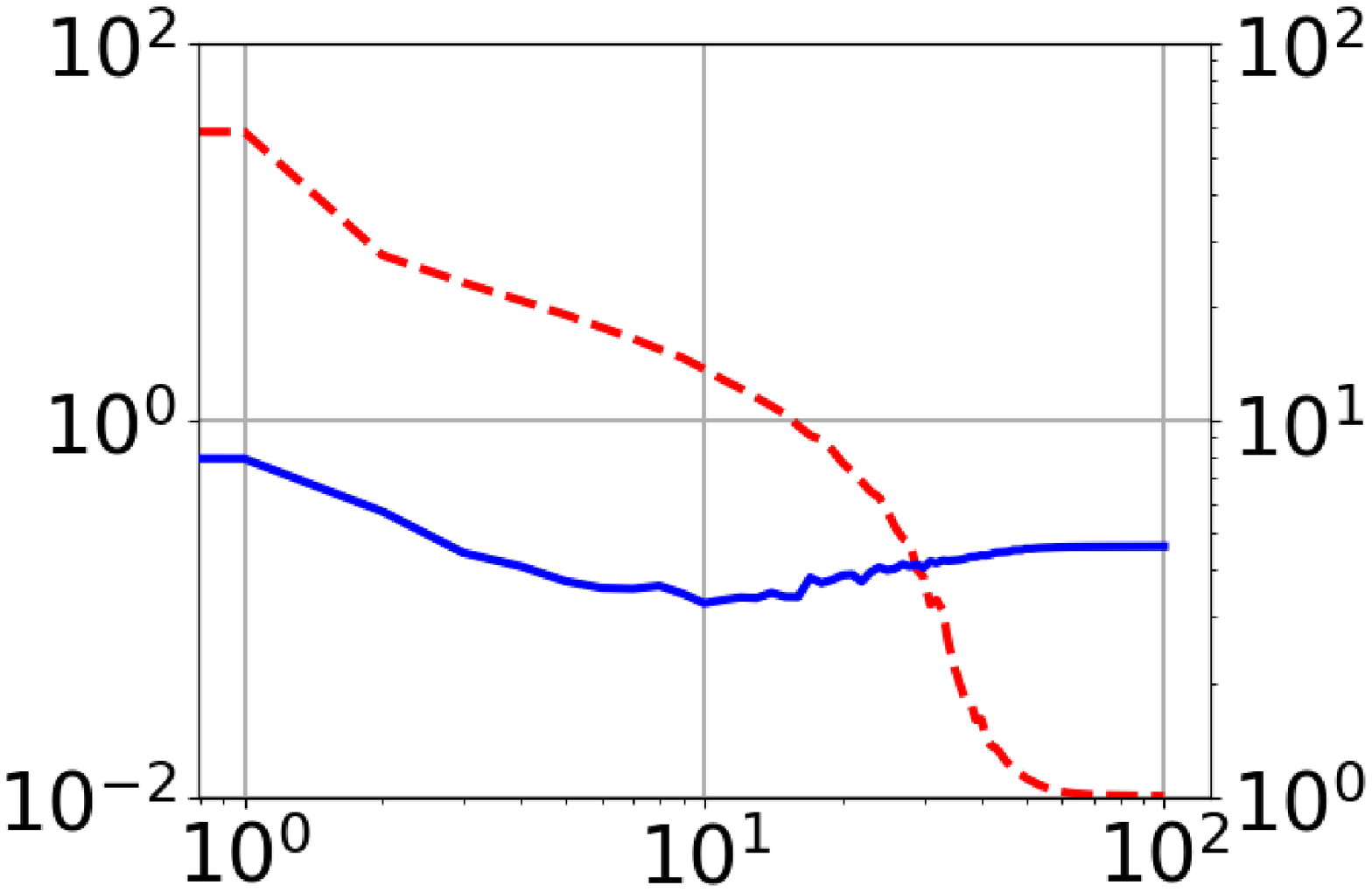} &
\includegraphics[width=\fw]{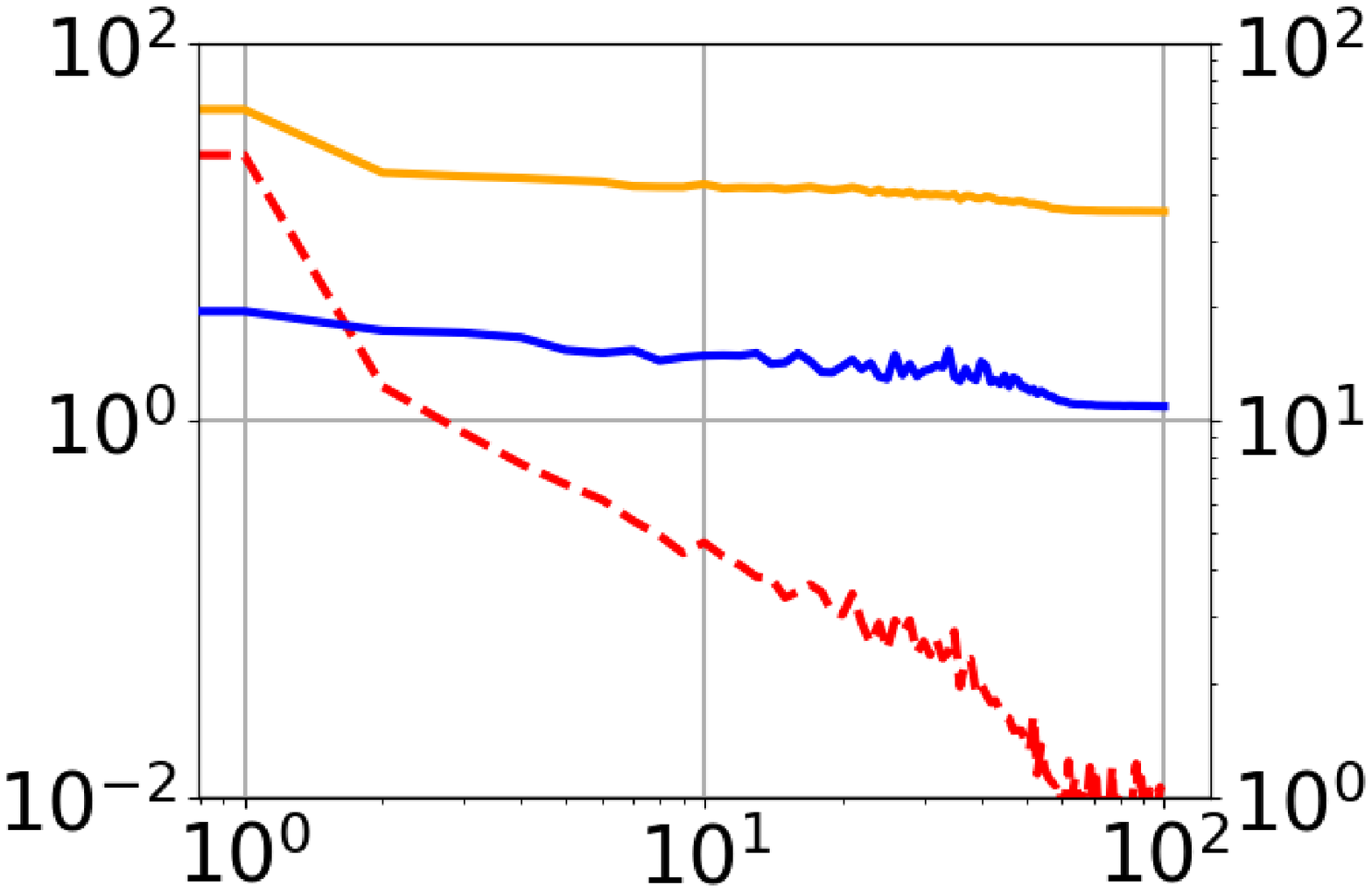} \\
\quad $\tr=0$ (baseline) & 
\quad $\tr=10\%$ & 
&
\quad $\tr=0$ & 
\quad $\tr=10\%$ \\
\multicolumn{2}{c}{\small \quad (1) NN-2} && \multicolumn{2}{c}{\small \quad (2) LeNet}\\
\end{tabular}
\vspace{-5pt}
\caption{[Effect of example-trimming] Training loss of the mini-batch examples (orange line with 1st y-axis) and the trimmed examples with fixed $\tr=10\%$ (red dotted-line with 1st y-axis), and the test loss (blue solid-line with 2nd y-axis) over epoch (x-axis) for MNIST by NN-2 network (left) and LeNet (right) trained using SGD with sigmoid learning-rate: The label noise is not applied.}
\label{fig:pretest:trimming}
\end{figure}
%
%
%
%
%
%
%
\def\fw{80pt}
\def\pw{70pt}
\begin{figure} [htb]
\centering
\scriptsize
\begin{tabular}{P{\pw}P{\pw}P{\pw}P{\pw}P{\pw}P{\pw}P{\pw}}
\includegraphics[width=\fw]{ep-lossm,MNIST-sigmoid-0.01-0.0001-1.0,FFNet_MNIST,e,100,B,128,nB,1,dB,1,M,1,TrimData0.00,1.0,0.0,0.0,0.0,mo,0.9,TrimData} &
\includegraphics[width=\fw]{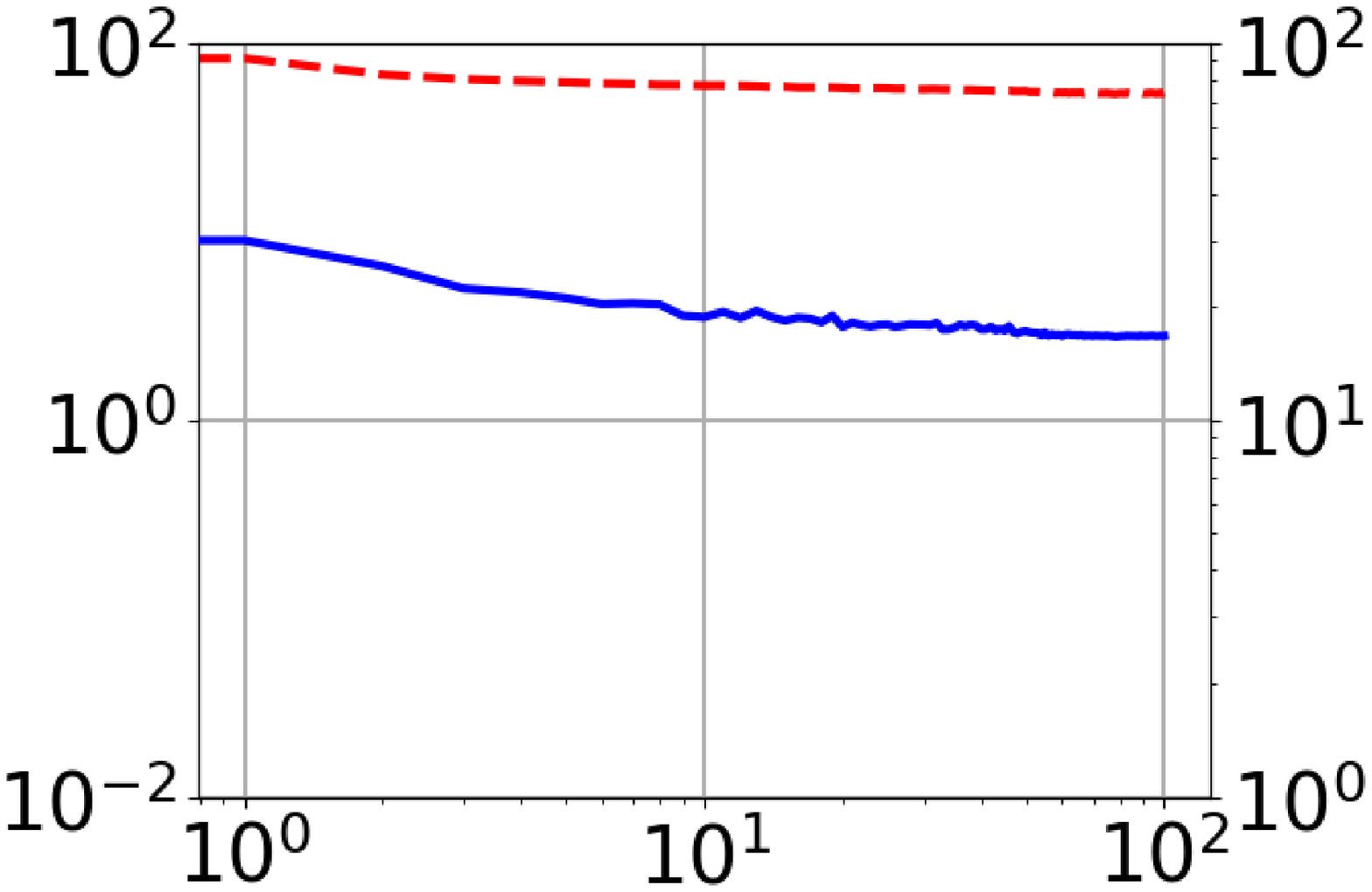} & 
\includegraphics[width=\fw]{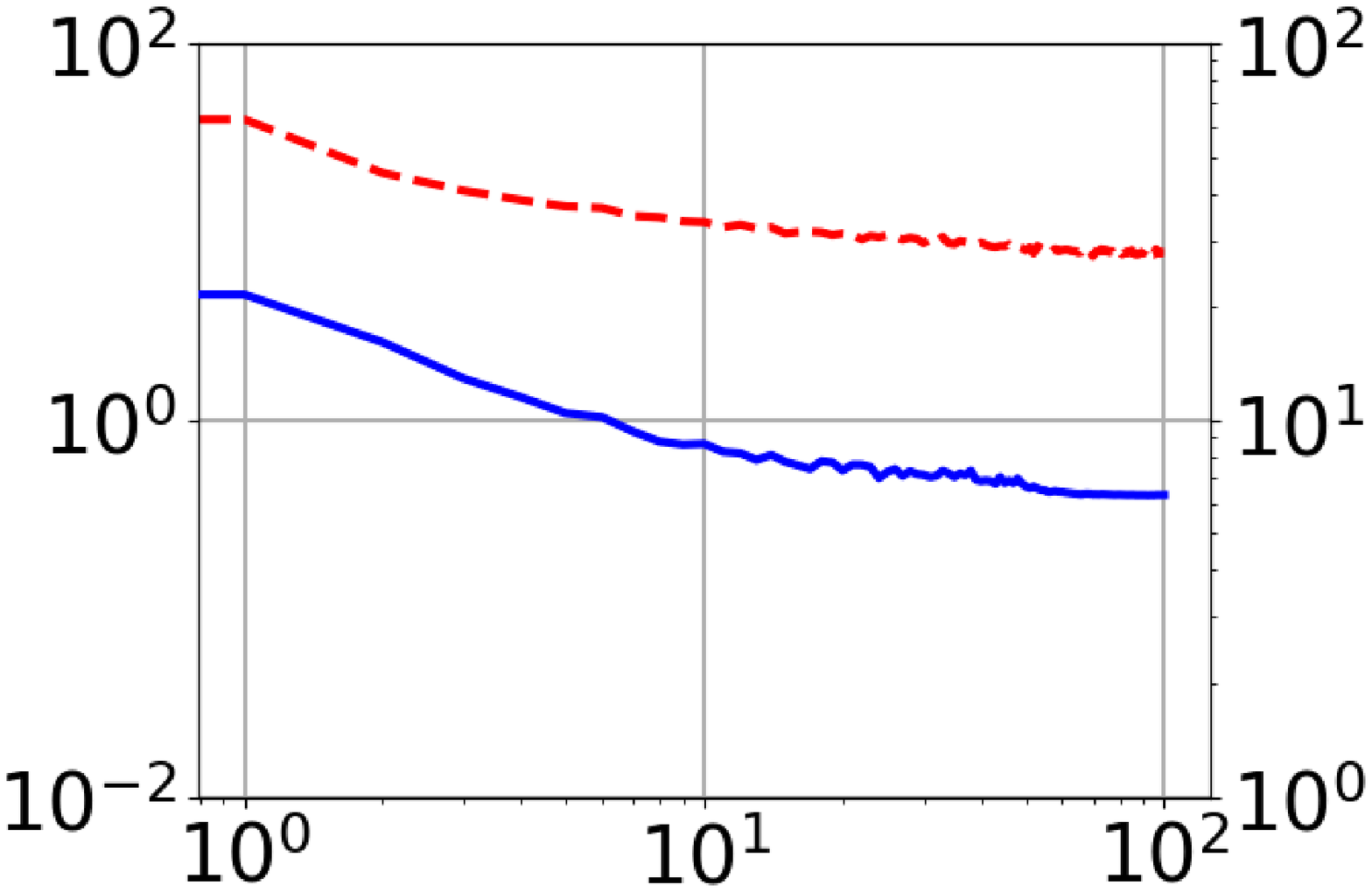} & 
\includegraphics[width=\fw]{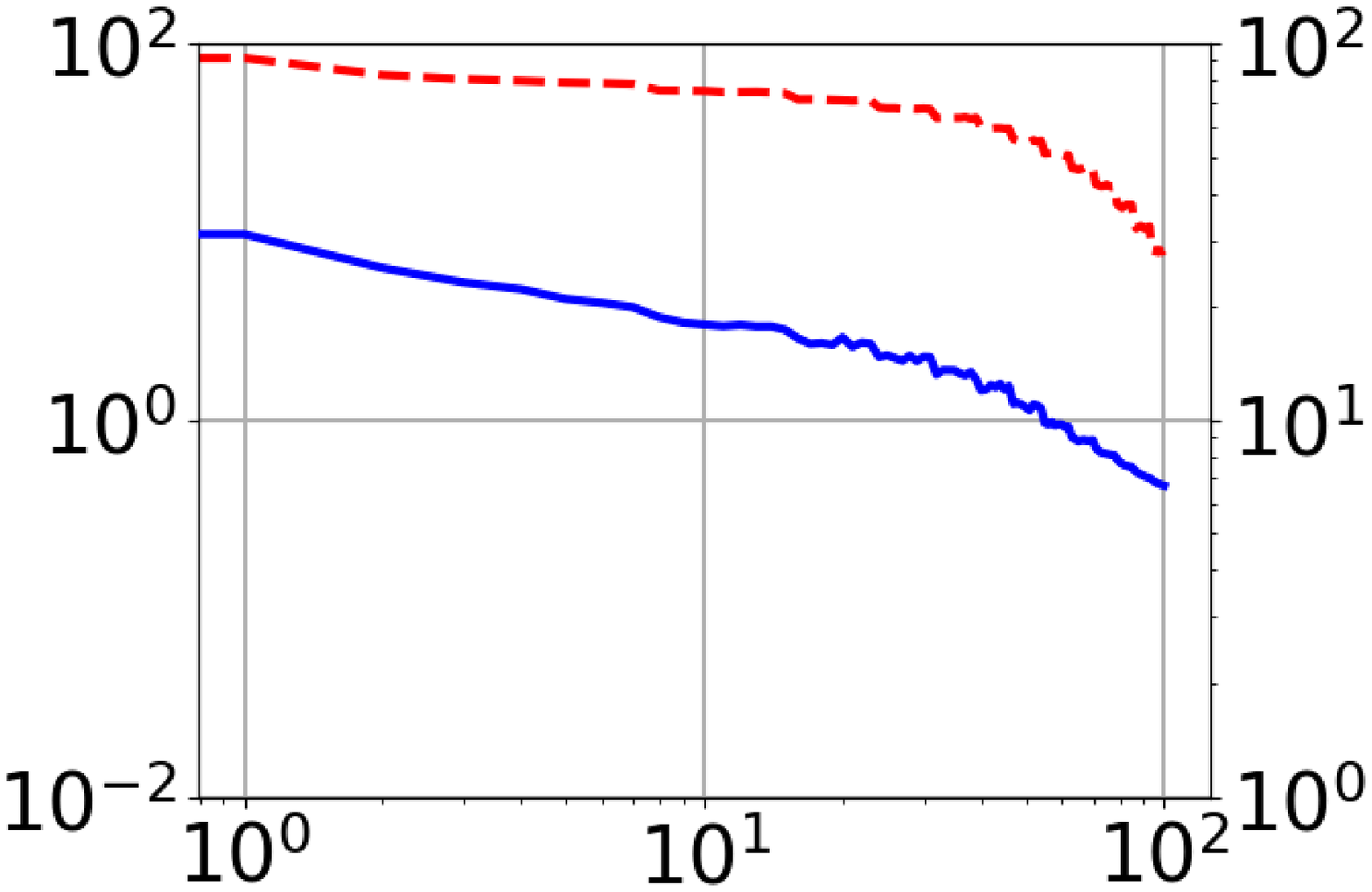} \\
\includegraphics[width=\fw]{ep-lossm,MNIST-sigmoid-0.01-0.0001-1.0,ConvNet_MNIST,e,100,B,128,nB,1,dB,1,M,1,TrimData0.00,1.0,0.0,0.0,0.0,mo,0.9,TrimData} &
\includegraphics[width=\fw]{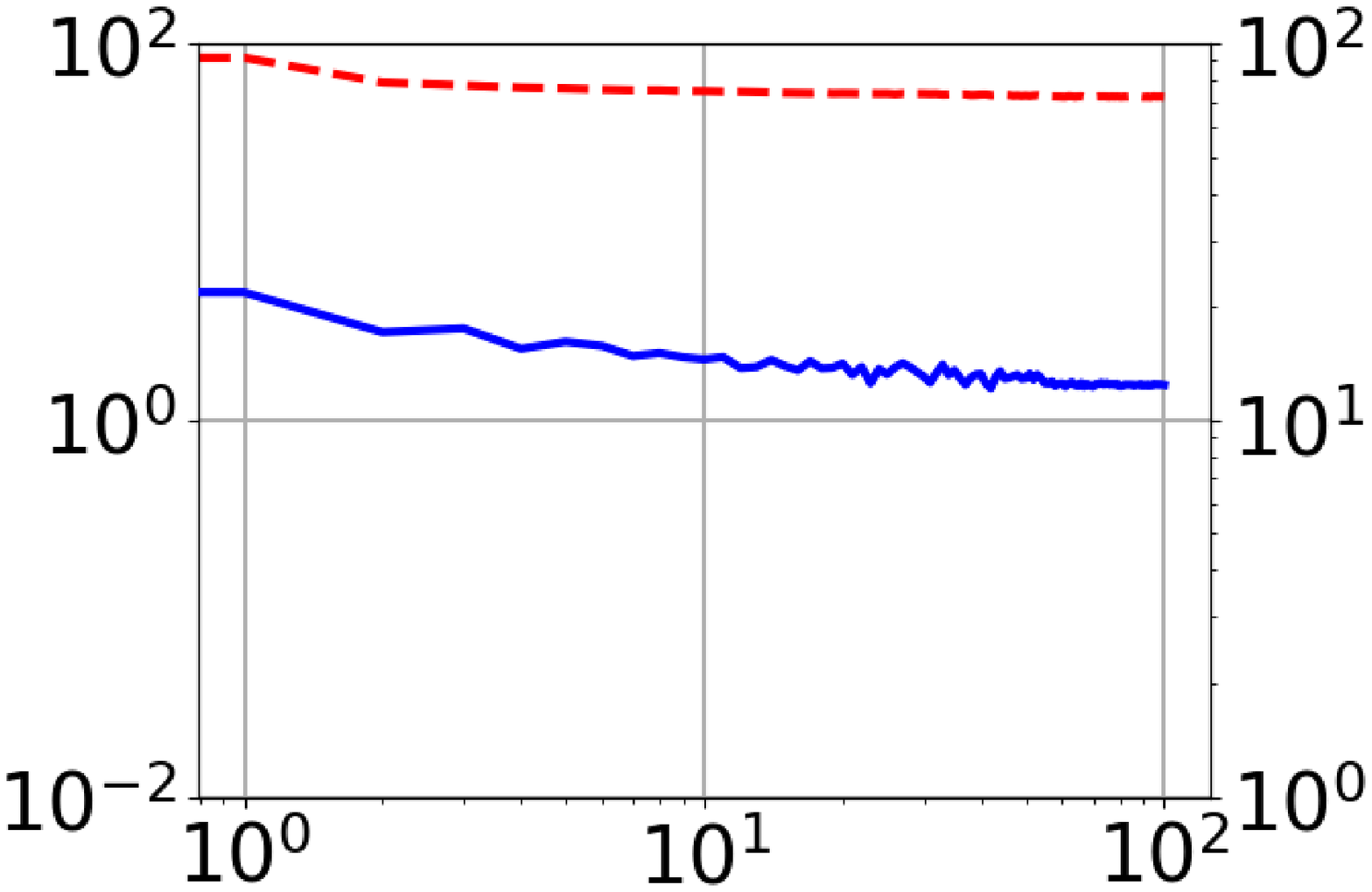} & 
\includegraphics[width=\fw]{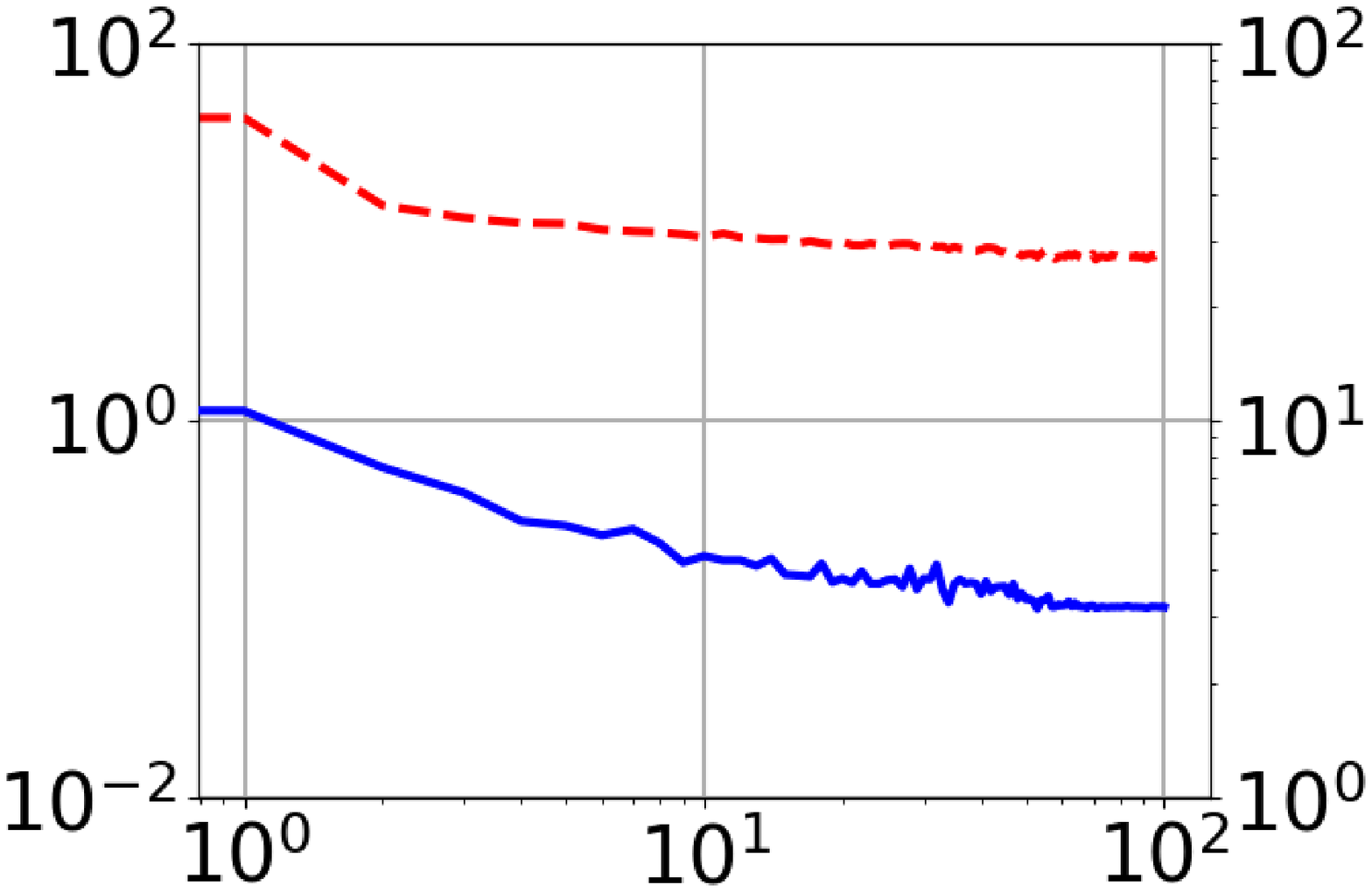} & 
\includegraphics[width=\fw]{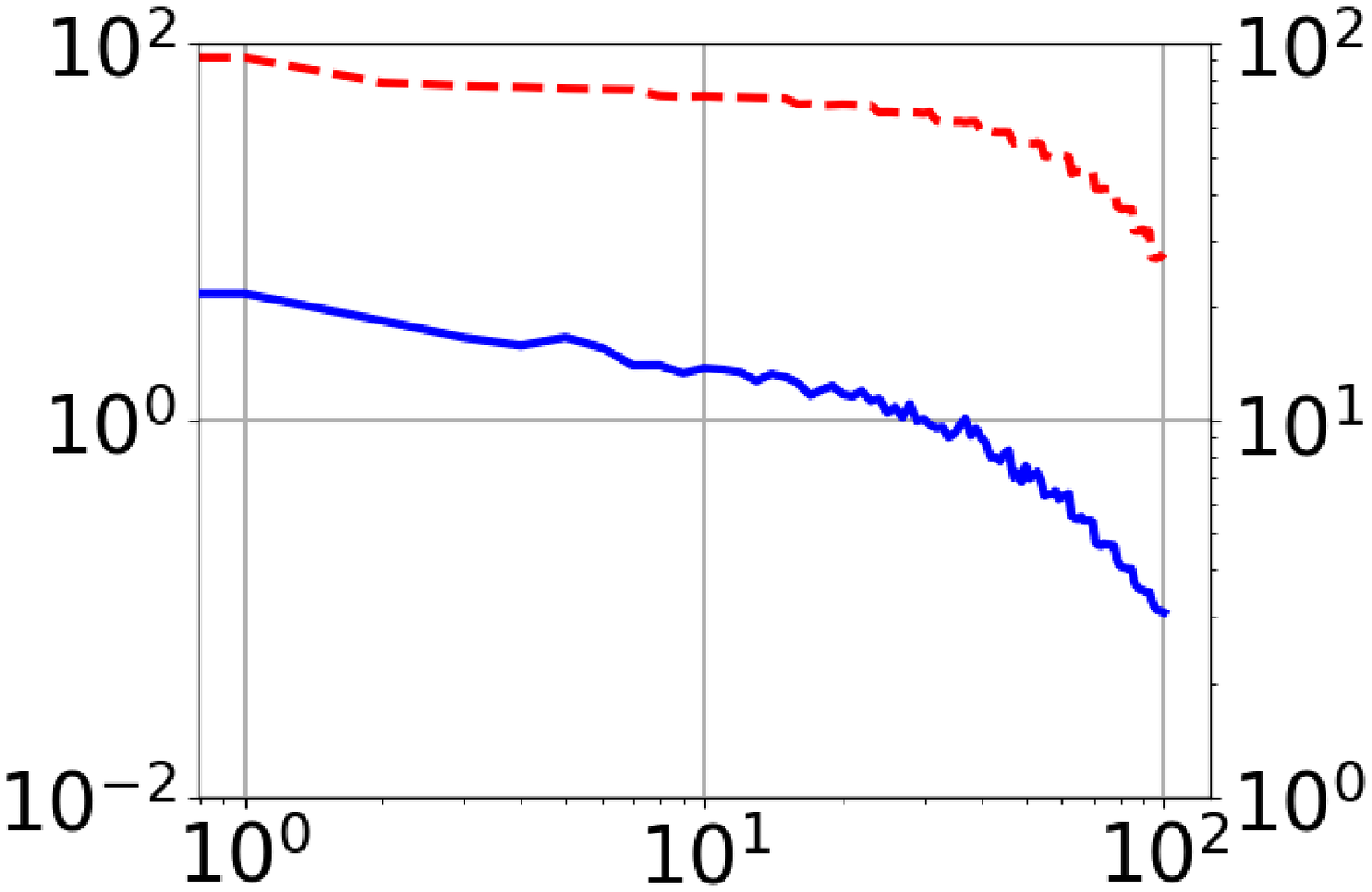} \\
\quad $\nr=0,\tr=0$ & 
\quad $\nr=10\%,\tr=0$ & 
$\nr=10\%,\tr=20\%$ &
$\nr=10\%,\tr=20\%$ \\
\quad baseline SGD & \quad label noise & \quad constant trimming  & \quad linear trimming \\
\end{tabular}
\vspace{-5pt}
\caption{[Combination of trimming with noise] Training loss of $\tr$-trimmed examples (red dotted-line with 1st y-axis) and the test loss (blue solid-line with 2nd y-axis) over epoch (x-axis) for MNIST by NN-2 network (top) and LeNet (bottom) trained using (1st, ..., 4th columns) the baseline SGD, the label-noise, a constant trimming with the noise, and a linear trimming with the noise.}
\label{fig:experiment:ours:process}
\end{figure}
\def\fw{95pt}
\def\pw{95pt}
\begin{figure} [htb]
\centering
\scriptsize

\begin{tabular}{c m{\pw}m{\pw}m{\pw}}
& \multicolumn{3}{c}{\includegraphics[width=240pt]{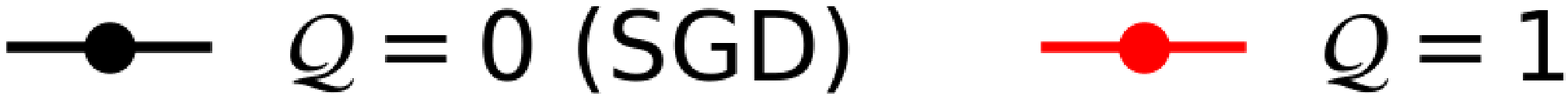}} \\
\parbox[t]{2mm}{\multirow{-2.2}{*}{\rotatebox[origin=c]{90}{   MNIST  }}} &
\includegraphics[width=\fw]{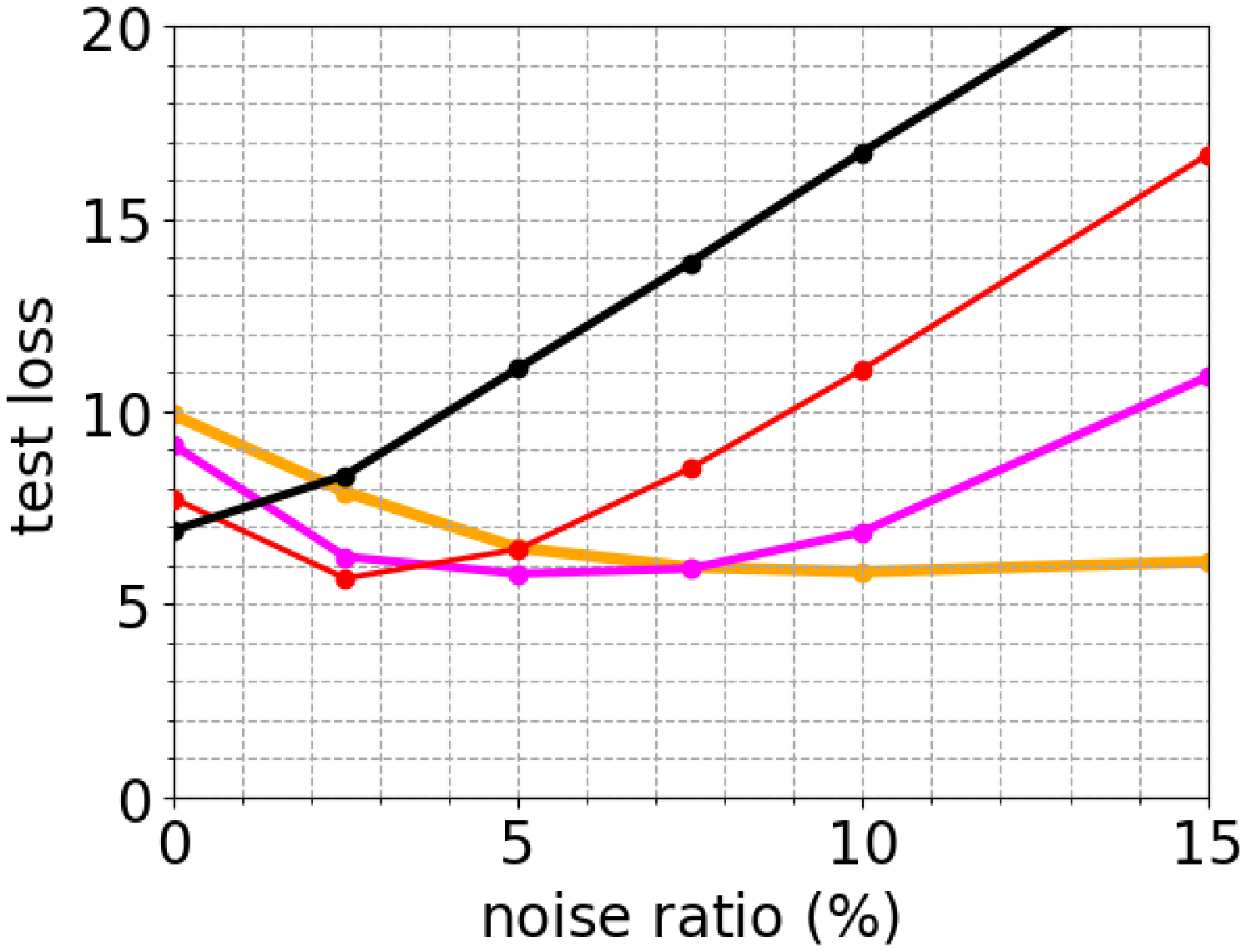} &
\includegraphics[width=\fw]{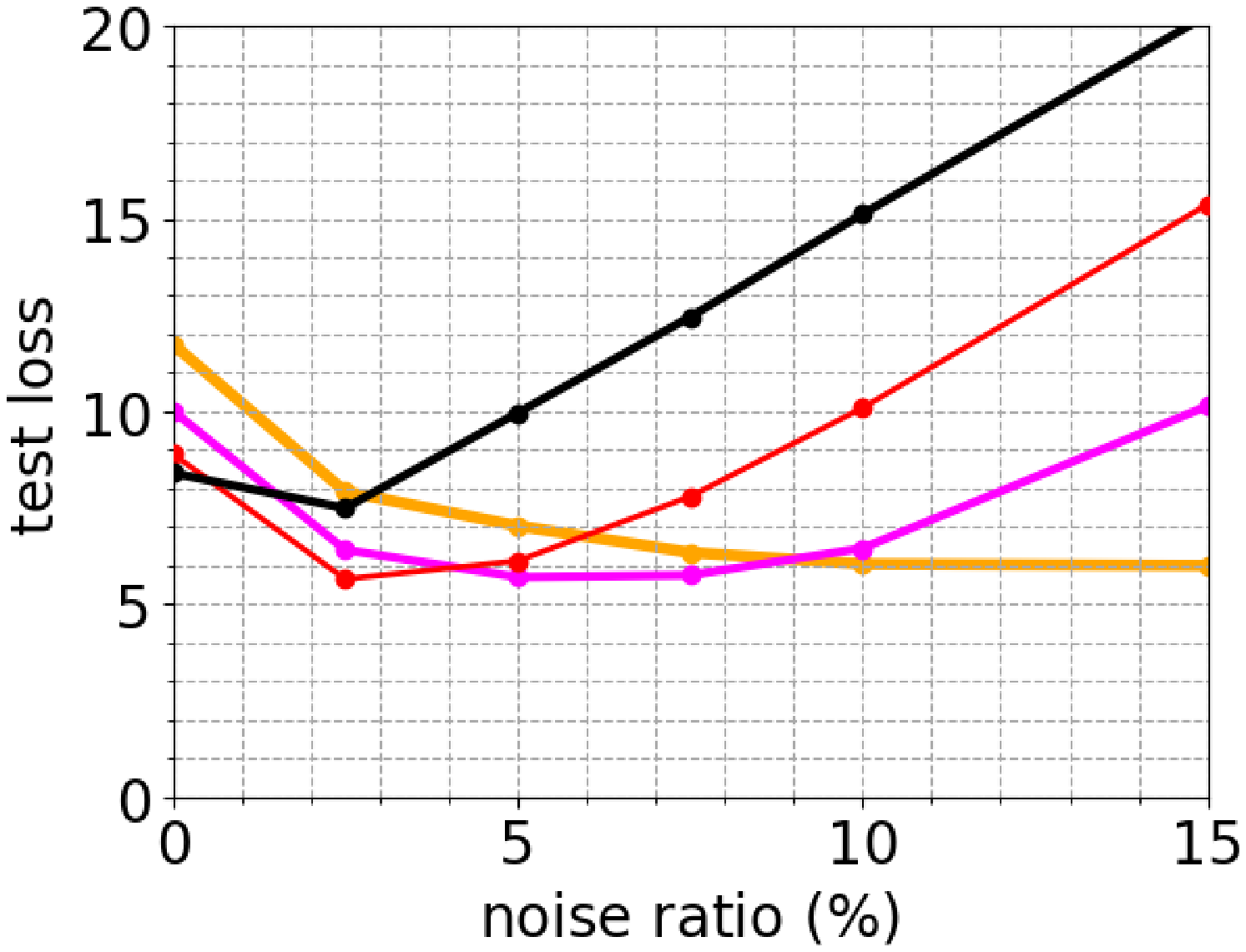} &
\includegraphics[width=\fw]{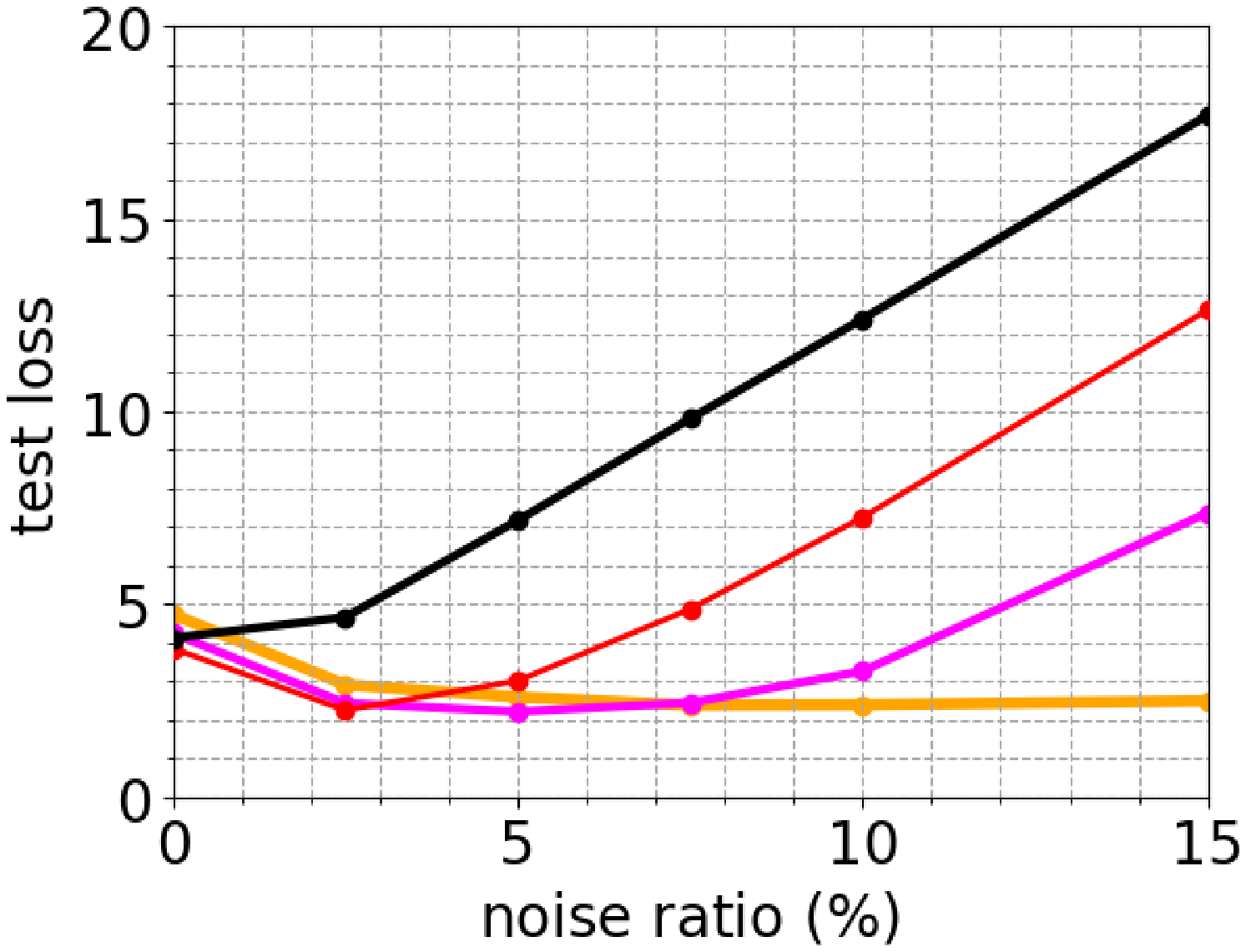} \\
\parbox[t]{2mm}{\multirow{-3.2}{*}{\rotatebox[origin=c]{90}{  Fashion-MNIST  }}} &
\includegraphics[width=\fw]{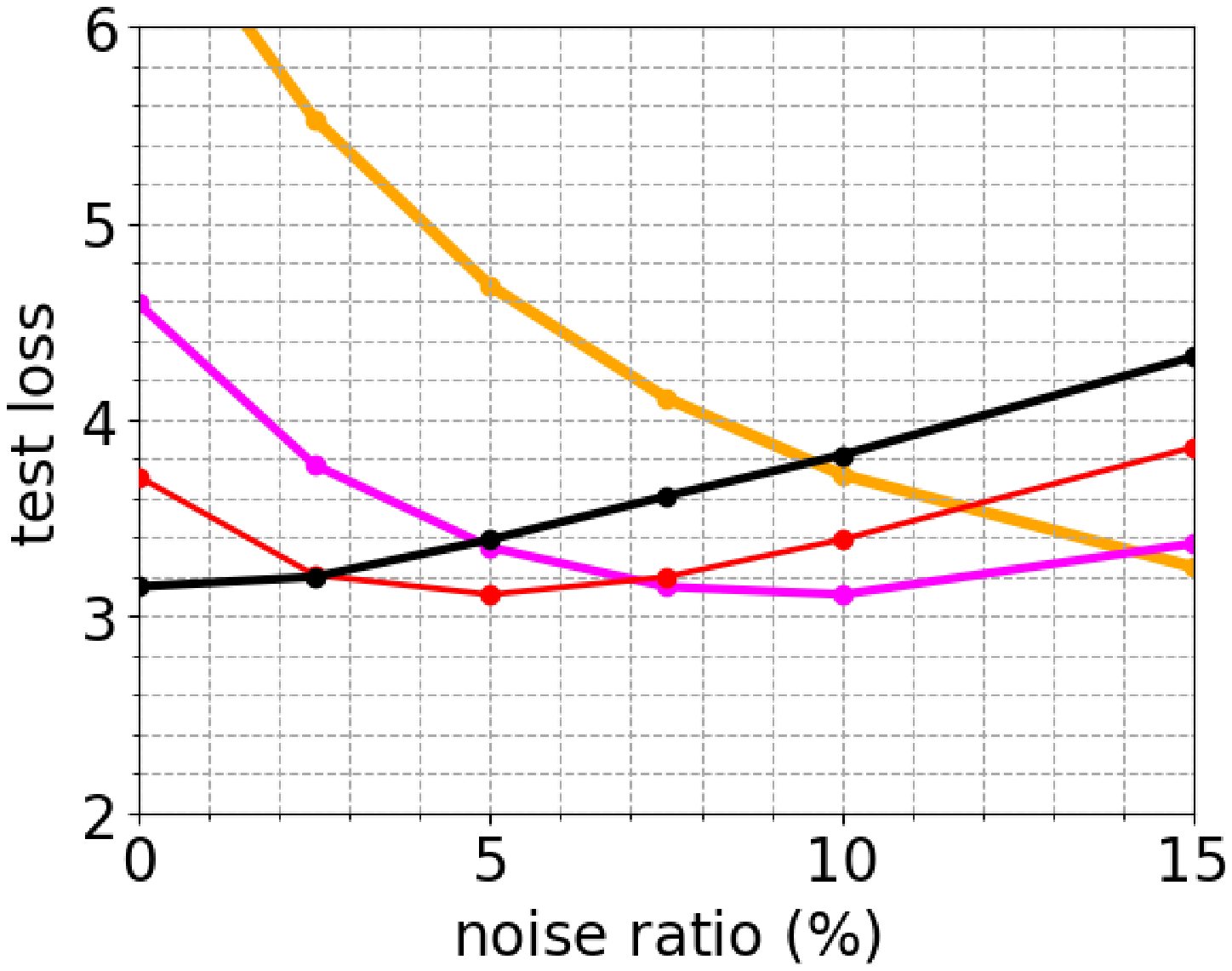} &
\includegraphics[width=\fw]{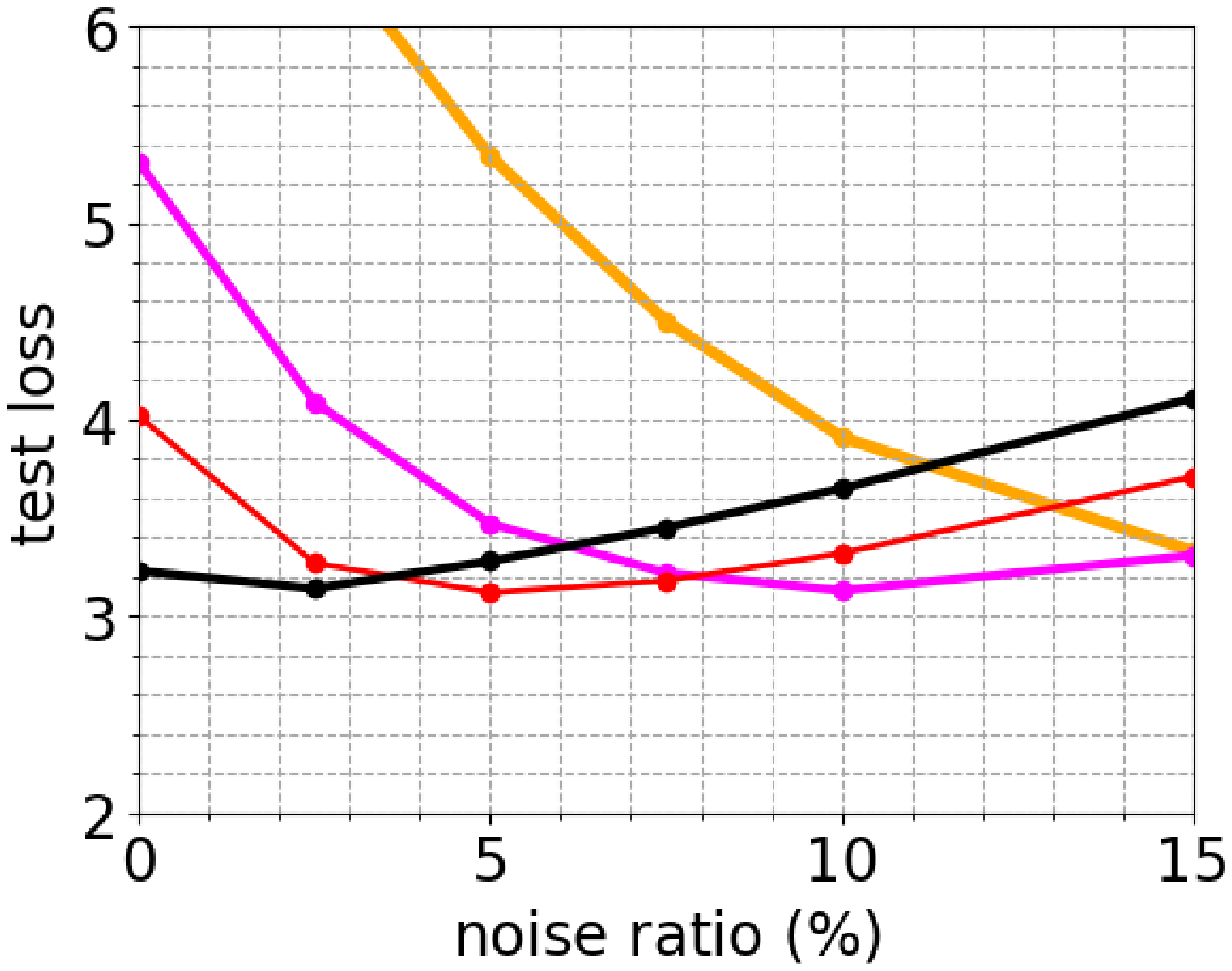} &
\includegraphics[width=\fw]{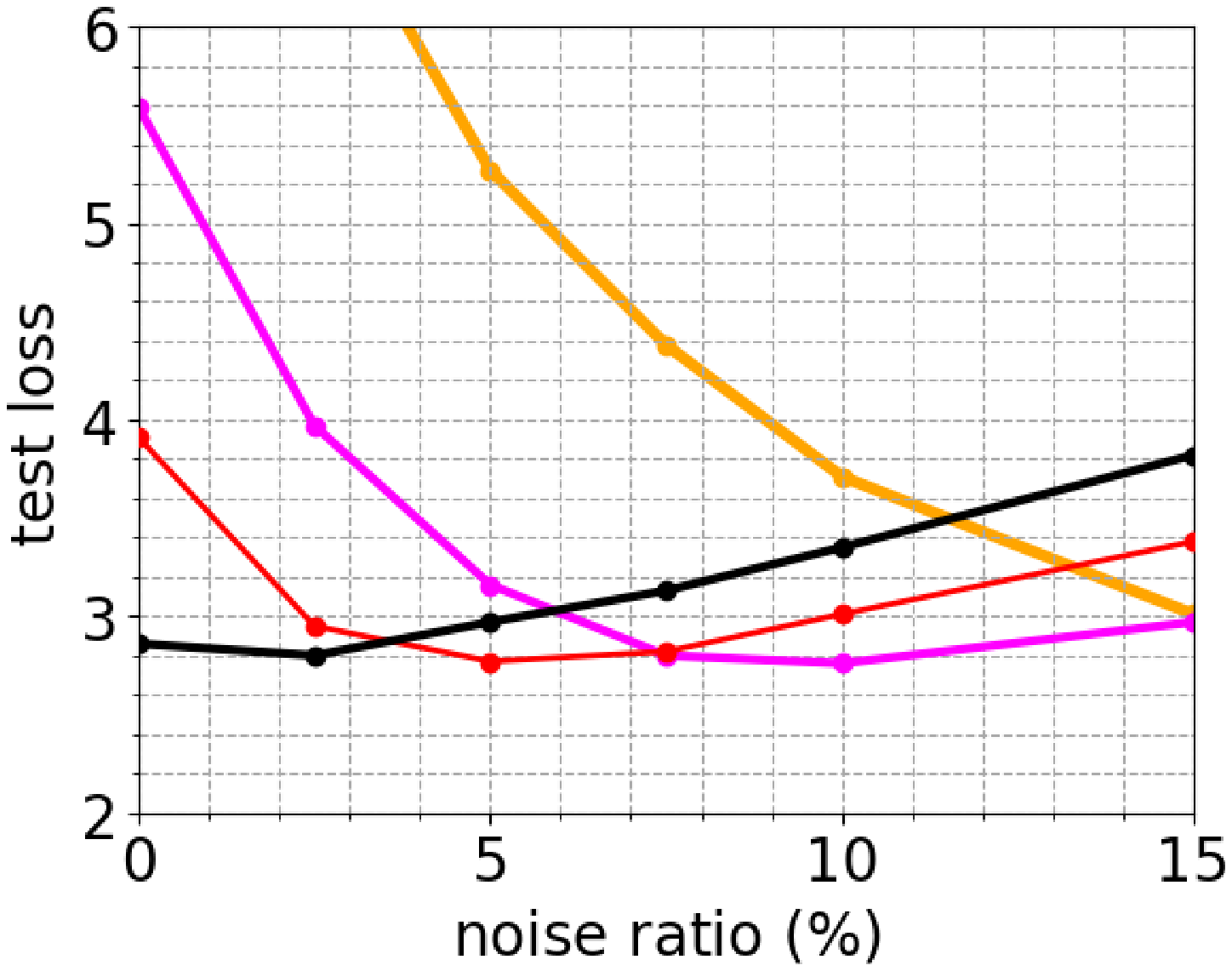} \\
\parbox[t]{2mm}{\multirow{-3.2}{*}{\rotatebox[origin=c]{90}{ EMNIST-Letters  }}} &
\includegraphics[width=\fw]{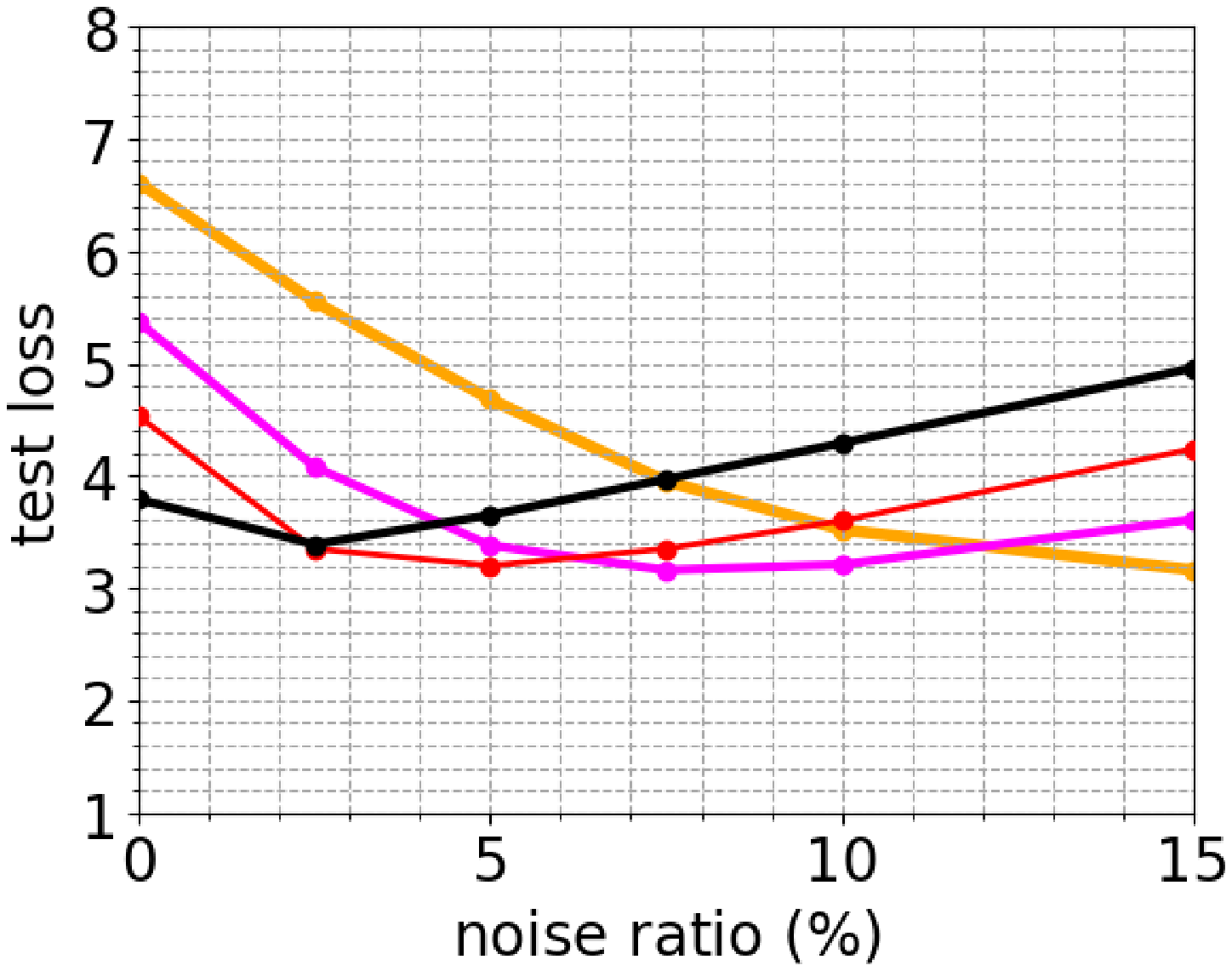} &
\includegraphics[width=\fw]{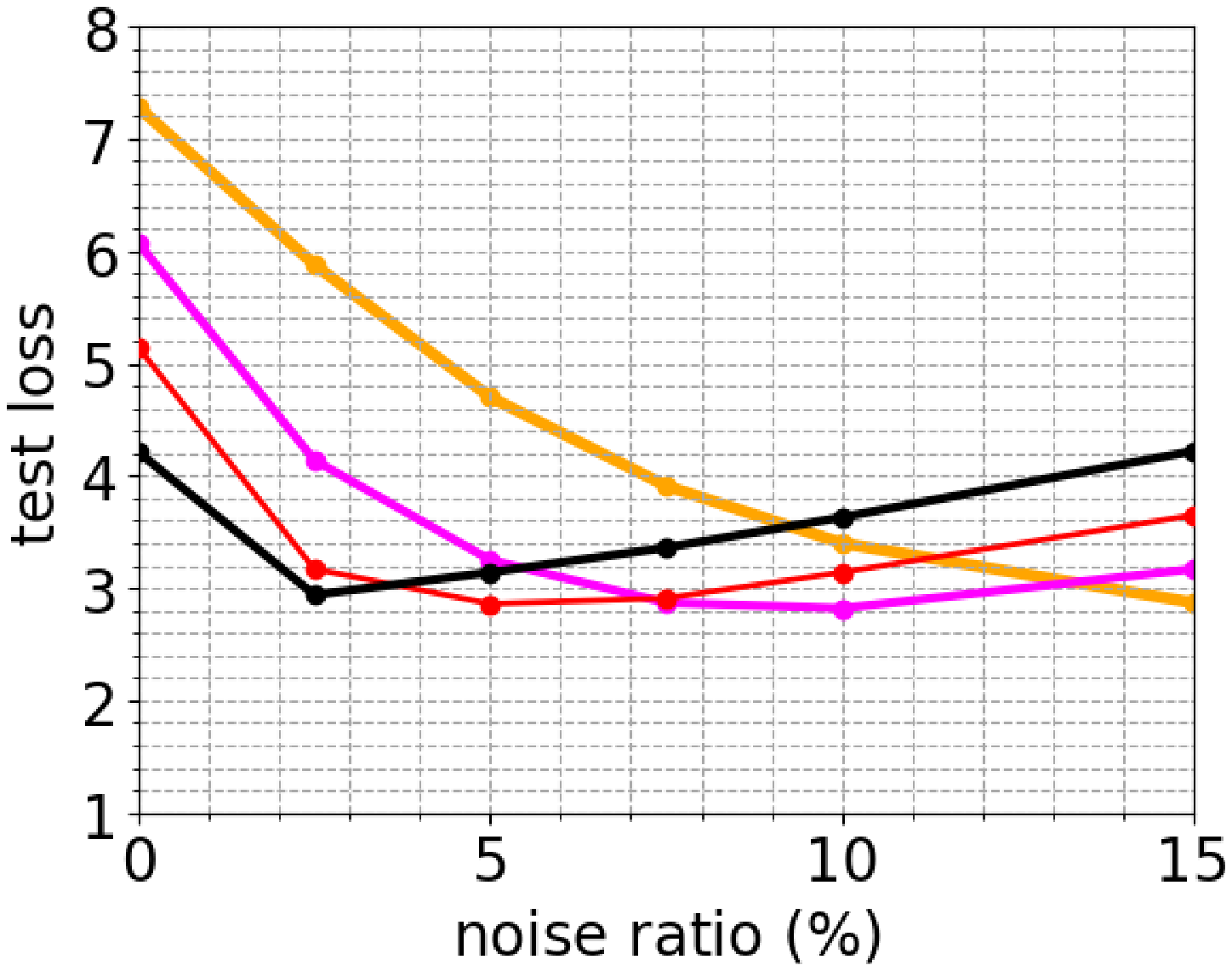} &
\includegraphics[width=\fw]{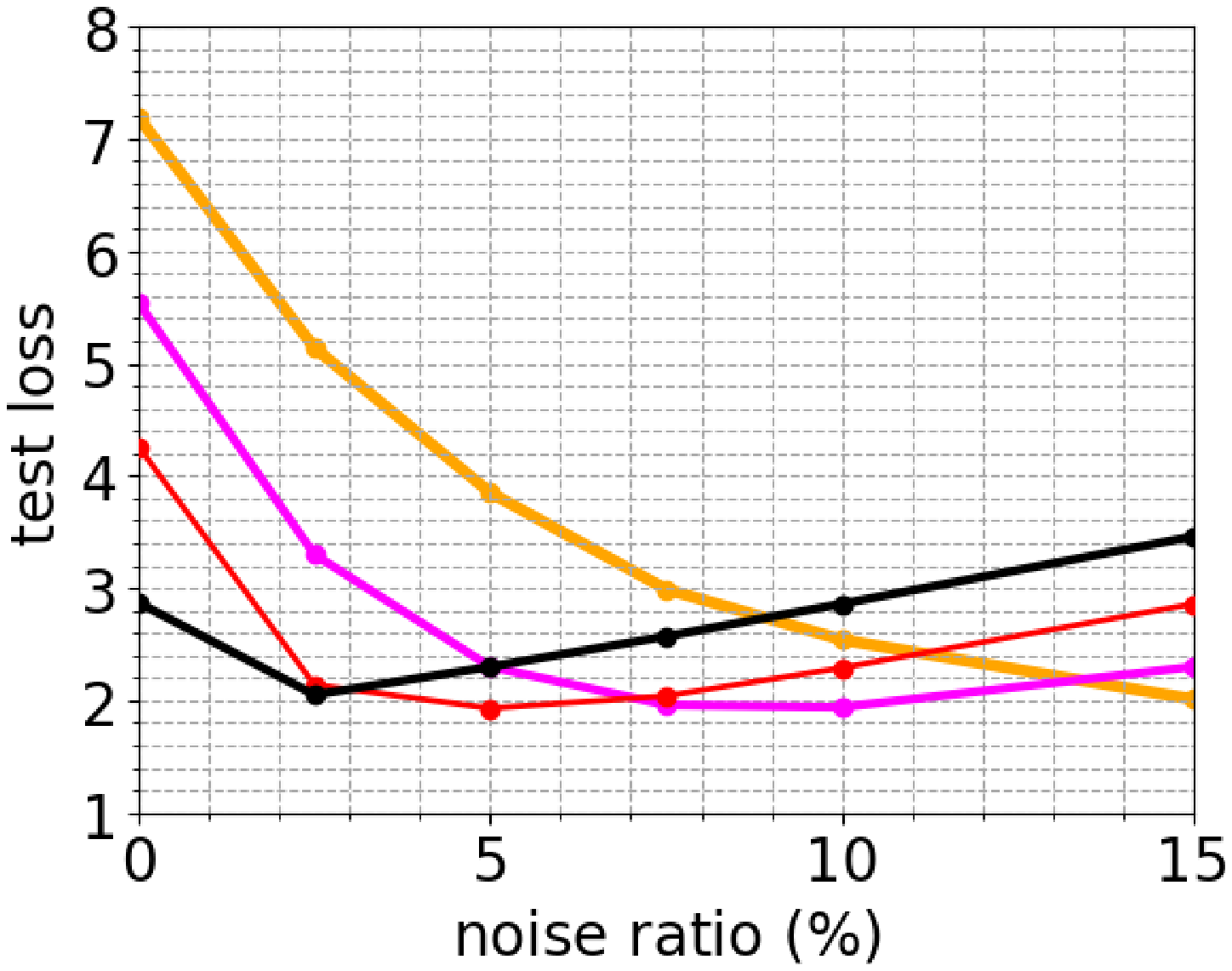} \\
& \multicolumn{1}{c}{NN-2} & \multicolumn{1}{c}{NN-3} & \multicolumn{1}{c}{LeNet} \\
\end{tabular}
\vspace{-10pt}
\caption{
Test loss (y-axis) over the label noise $\nr$ (x-axis) for MNIST (top), Fashion-MNIST (middle), and EMNIST-Letters (bottom) by (1st, 2nd, 3rd columns) NN-2, NN-3, LeNet 
optimized using 
SGD (black line), 
and ours with $\tr=10\%$ (red thin-line),  $\tr=20\%$ (magenta line), and $\tr=40\%$ (orange bold-line):
the test loss is averaged within 10 individual trials.}
\label{fig:experiment:ours:loss}
\end{figure}
%
%
%

%
%
\par
\vspace{3pt}
\noindent {\bf Effect of example trimming: } 
We also re-examine the example-trimming algorithm that prunes the top-$\tr/2$ and the bottom-$\tr/2$ examples using training loss with constant $\tr$ without the label noise.
Figure~\ref{fig:pretest:trimming} presents instances of 
the loss of training examples in the mini-batch (orange line), 
the loss of trimmed examples (red dotted line) that is used in the model update, 
and the test loss (blue solid-line) over epoch. 
As shown in Figure~\ref{fig:pretest:trimming}, 
the training with the example-trimming leads the model to a local minima where both the loss for the whole training examples (orange line) and the test examples (blue) are larger than the original SGD.
It is also observed that the example trimming slows the convergence of the test loss as the label noise.
\subsection{Experiment on Label-Noised Trim-SGD}
We experiment the proposed algorithm that combines the label noise with the example-trimming.
Figure~\ref{fig:experiment:ours:process} presents the training loss of trimmed examples (red dotted-line) 
and the test loss (blue line) over epoch for MNIST by NN-2 and LeNet, optimized using
(1st column) the baseline SGD, 
(2nd column) SGD using the label noise of $\nr=10\%$,
(3rd column) ours using $\nr=10\%$ with fixed $\tr=20\%$,
and (4th column) ours using a linear trimming with $\tr=20\%$, respectively.
The constant trimming (3rd column in Figure~\ref{fig:experiment:ours:process}) is the vanilla implementation of our Label-Noised Trim-SGD and is shown to successfully improves the test loss.
We also propose a trimming ratio (4th column) as our practical implementation in which the trimming ratio increases as
\begin{eqnarray} \label{eq:linear-trimming}
\tr^{(\theta)} \coloneqq \theta \cdot \tr,
\end{eqnarray}
where $\theta \in [0,1]$ is the time evolution.  We use Eq.(\ref{eq:linear-trimming}) in the following experiments.
%
%
%
%

%
%
%
%
\def\pw{9mm}
\begin{table}[h!tb]
\caption{Test loss for MNIST, Fashion-MNIST, and EMNIST-Letters by NN-2, NN-3, LeNet 
optimized using (3rd, ..., 8th columns) SGD, RMSprop, Adam, Entropy-SGD, Accelerated-SGD, and ours with $\tr=20\%$: 
the sigmoid learning-rate annealing and momentum of 0.9 are employed,
(top part) test loss averaged within individual trials, and (bottom part) minimum test loss over iterations are presented.
}
\label{tab:comparion_to_others:loss}
\small
\centering
{\small (1) Mean test loss}\\
\begin{tabular}{l | P{\pw} | P{\pw}P{\pw}P{\pw}P{\pw}P{\pw}P{\pw}}
\hline
data-set    & model & SGD  & RMS  & Adam & eSGD & aSGD & Ours \\ 
\hline
MNIST          & NN-2  & 6.91 & 8.28 & 8.26 & 6.29 & 6.84 & \textbf{5.67} \\
MNIST          & NN-3  & 7.49 & 8.19 & 8.14 & 6.39 & 7.53 & \textbf{5.65} \\
MNIST          & LeNet & 4.12 & 3.77 & 3.85 & 2.72 & 3.66 & \textbf{2.25} \\
\hline
Fashion-MNIST  & NN-2  & 3.15 & 3.37 & 3.36 & 3.45 & 3.34 & \textbf{3.11} \\
Fashion-MNIST  & NN-3  & 3.14 & 3.35 & 3.38 & 3.30 & 3.29 & \textbf{3.12} \\
Fashion-MNIST  & LeNet & 2.80 & 3.01 & 2.93 & 2.93 & 2.85 & \textbf{2.77} \\
\hline
EMNIST-Letters & NN-2  & 3.38 & 3.57 & 3.56 & 3.53 & 3.42 & \textbf{3.16} \\
EMNIST-Letters & NN-3  & 2.95 & 3.19 & 3.18 & 3.09 & 2.99 & \textbf{2.82} \\
EMNIST-Letters & LeNet & 2.05 & 2.19 & 2.16 & 2.22 & 2.05 & \textbf{1.94} \\
\hline
\end{tabular}\\
\vspace{6pt}
{\small (2) Minimum test loss}\\
\begin{tabular}{l | P{\pw} | P{\pw}P{\pw}P{\pw}P{\pw}P{\pw}P{\pw}}
\hline
data-set    & model & SGD  & RMS  & Adam & eSGD & aSGD & Ours \\ 
\hline
MNIST          & NN-2  & 6.40 & 6.43 & 6.61 & 6.01 & 6.37 & \textbf{5.47} \\
MNIST          & NN-3  & 6.32 & 6.76 & 6.44 & 5.91 & 6.54 & \textbf{5.26} \\
MNIST          & LeNet & 2.54 & 2.63 & 2.68 & 2.42 & 2.83 & \textbf{1.89} \\
\hline
Fashion-MNIST  & NN-2  & 3.12 & 3.19 & 3.18 & 3.44 & 3.19 & \textbf{3.08} \\
Fashion-MNIST  & NN-3  & 3.09 & 3.11 & 3.17 & 3.27 & 3.17 & \textbf{3.08} \\
Fashion-MNIST  & LeNet & 2.74 & 2.71 & 2.66 & 2.84 & 2.70 & \textbf{2.62} \\
\hline
EMNIST-Letters & NN-2  & 3.35 & 3.43 & 3.51 & 3.48 & 3.35 & \textbf{3.10} \\
EMNIST-Letters & NN-3  & 2.87 & 2.99 & 3.04 & 2.96 & 2.94 & \textbf{2.71} \\
EMNIST-Letters & LeNet & 1.99 & 2.00 & 1.98 & 2.04 & 1.98 & \textbf{1.88} \\
\hline
\end{tabular}\\
\end{table}
%
%
%
%
%
\def\fw{100pt}
\def\pw{100pt}
\begin{figure} [h!tb]
\centering
\scriptsize
\includegraphics[width=300pt]{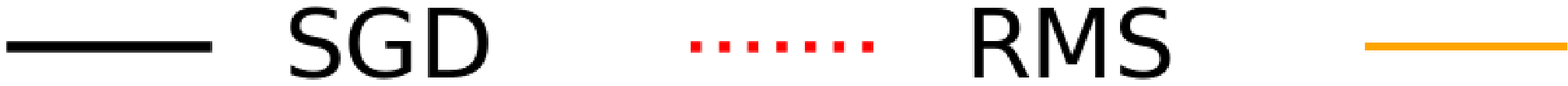}
\begin{tabular}{c m{\pw}m{\pw}m{\pw}}
\parbox[t]{2mm}{\multirow{-2.2}{*}{\rotatebox[origin=c]{90}{   MNIST  }}} &
\includegraphics[width=\fw]{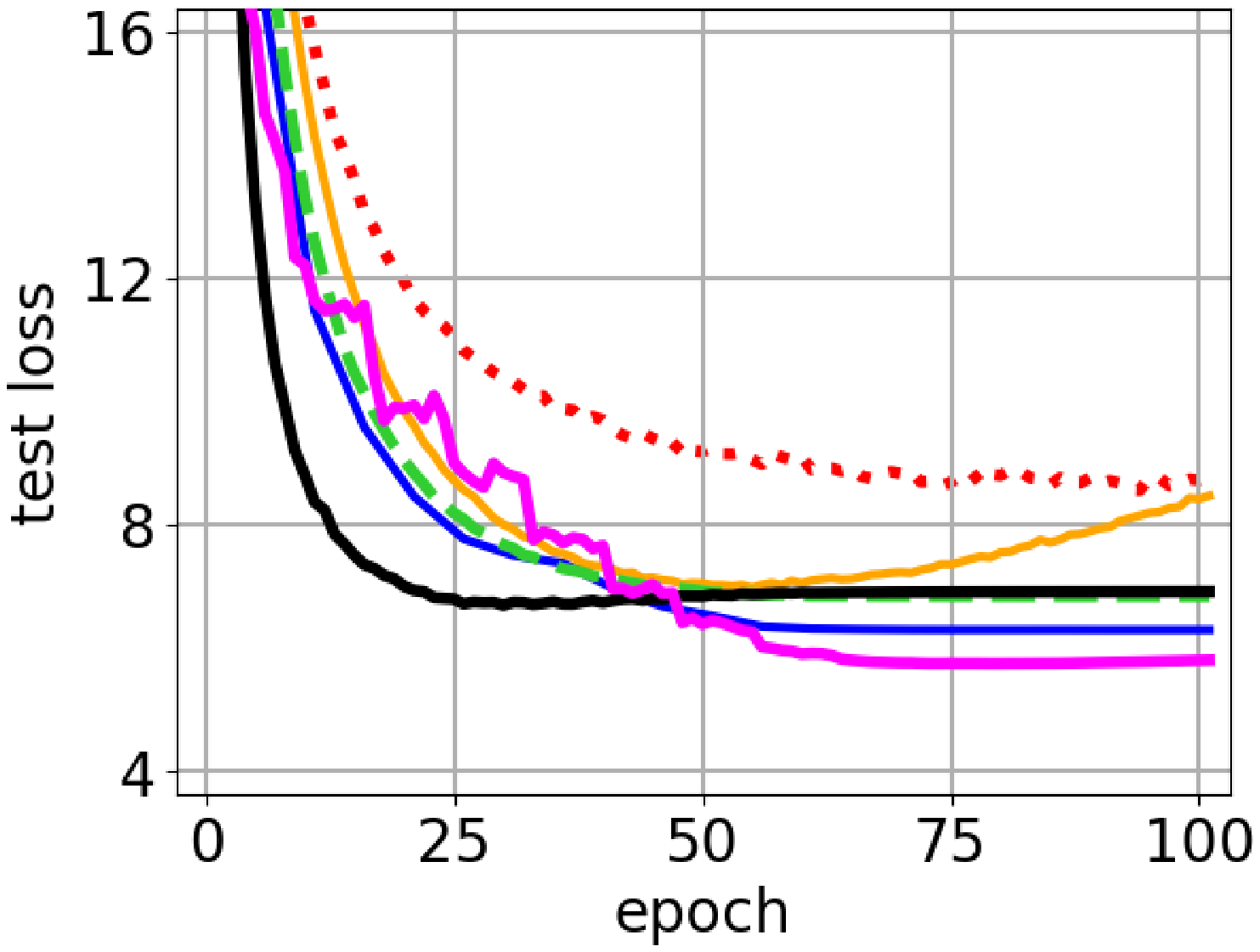} &
\includegraphics[width=\fw]{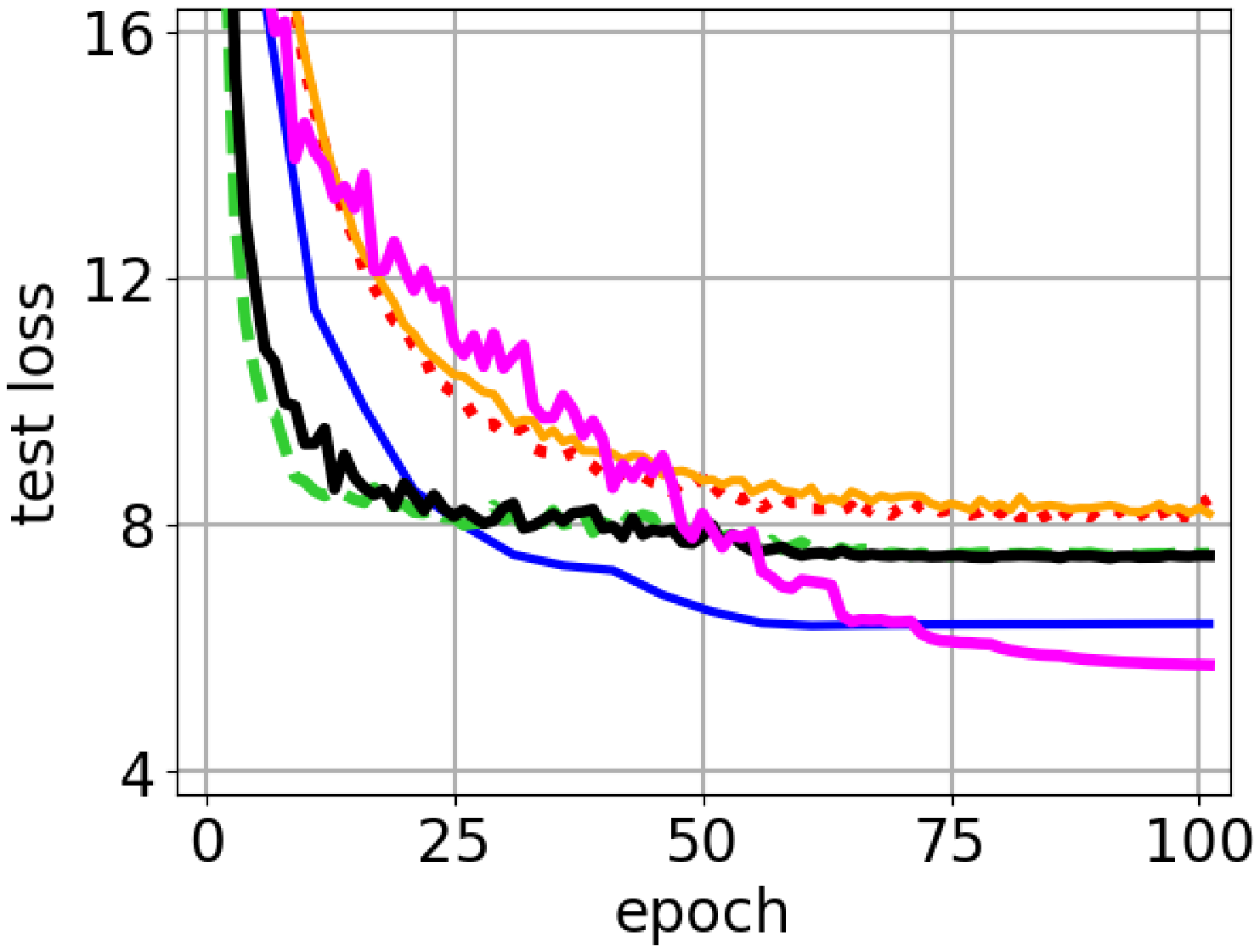} &
\includegraphics[width=\fw]{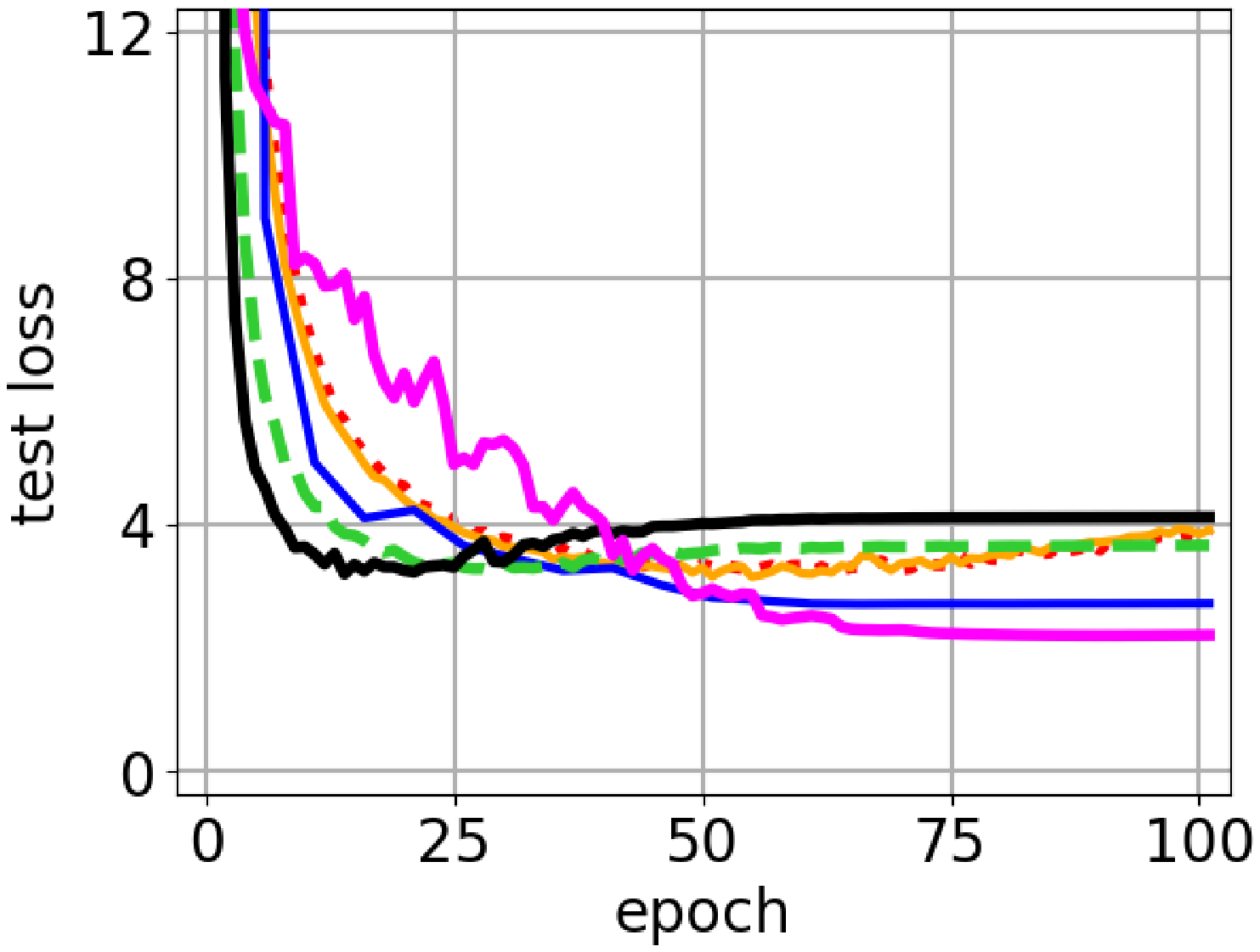} \\
\parbox[t]{2mm}{\multirow{-3.2}{*}{\rotatebox[origin=c]{90}{   Fashion-MNIST }}} &
\includegraphics[width=\fw]{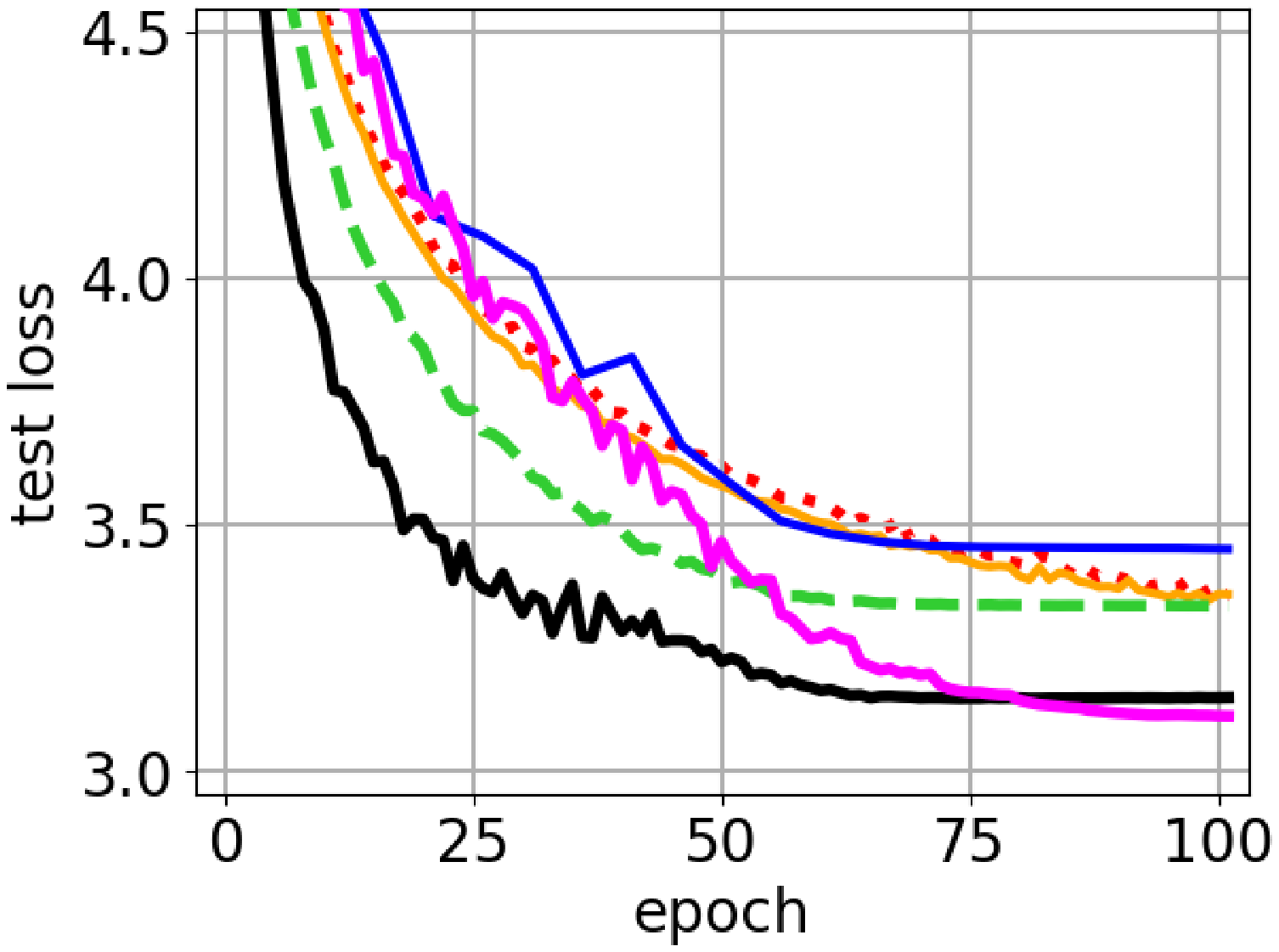} &
\includegraphics[width=\fw]{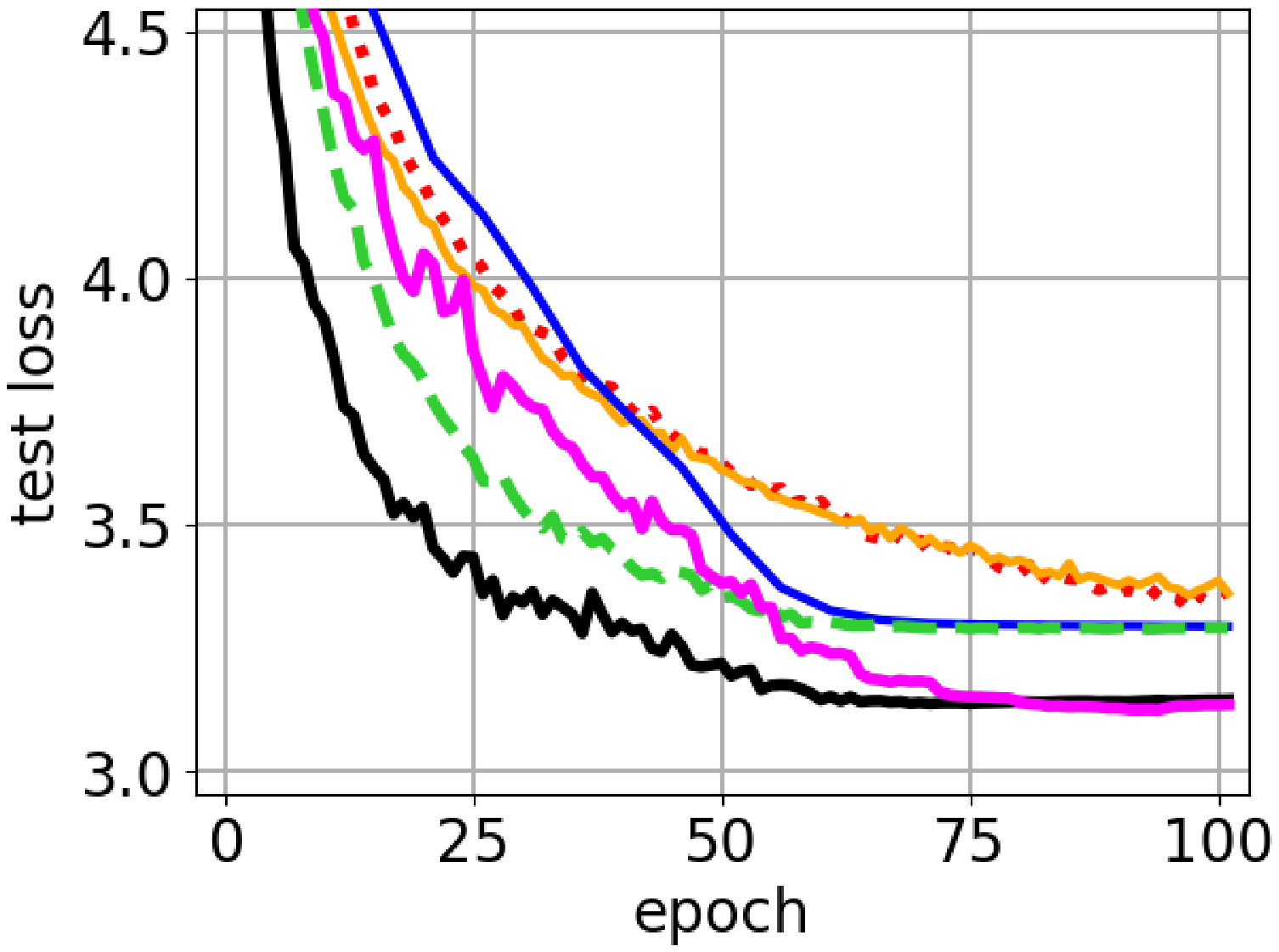} &
\includegraphics[width=\fw]{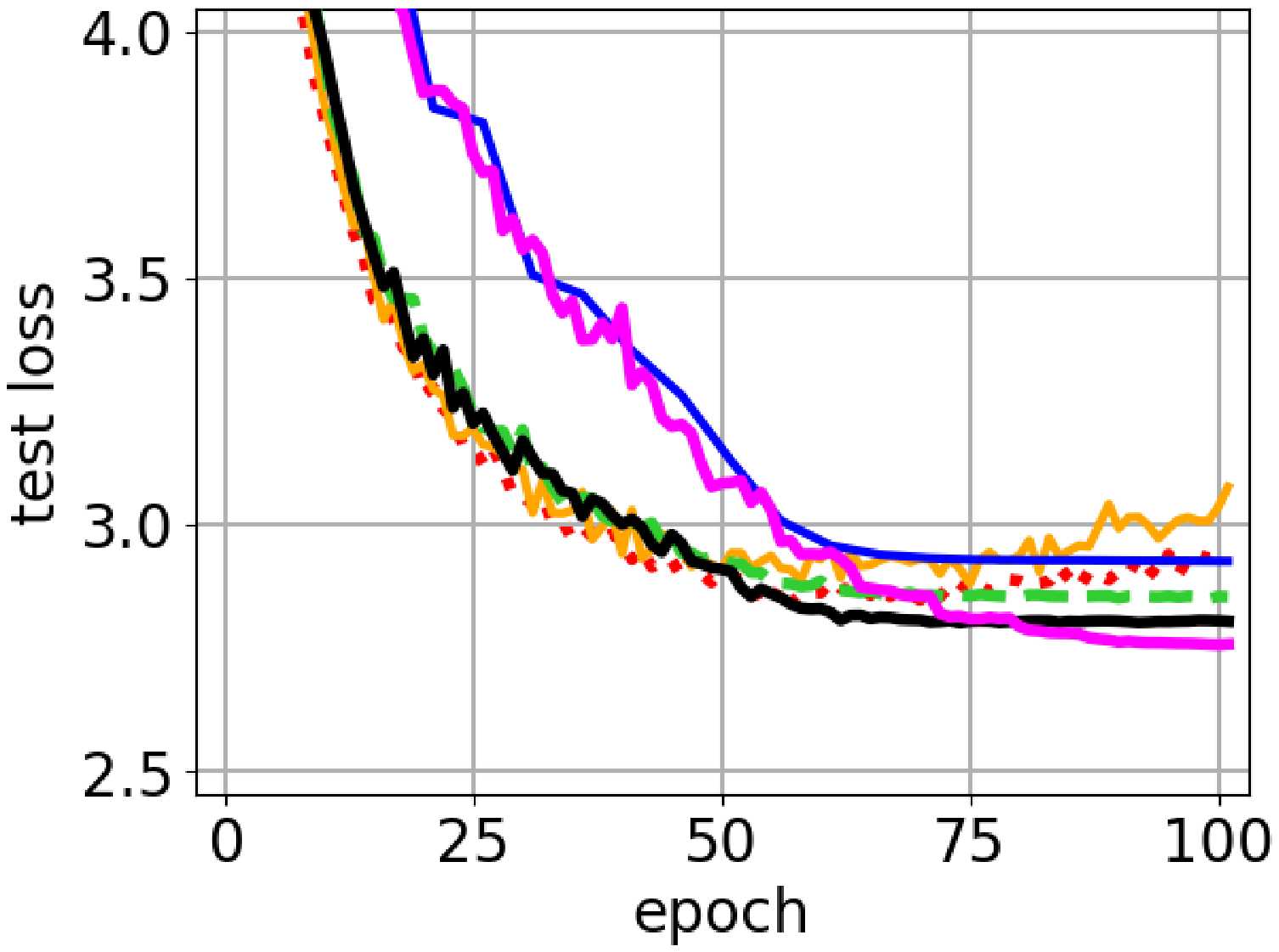} \\
\parbox[t]{2mm}{\multirow{-3.3}{*}{\rotatebox[origin=c]{90}{  EMNIST-Letters }}} &
\includegraphics[width=\fw]{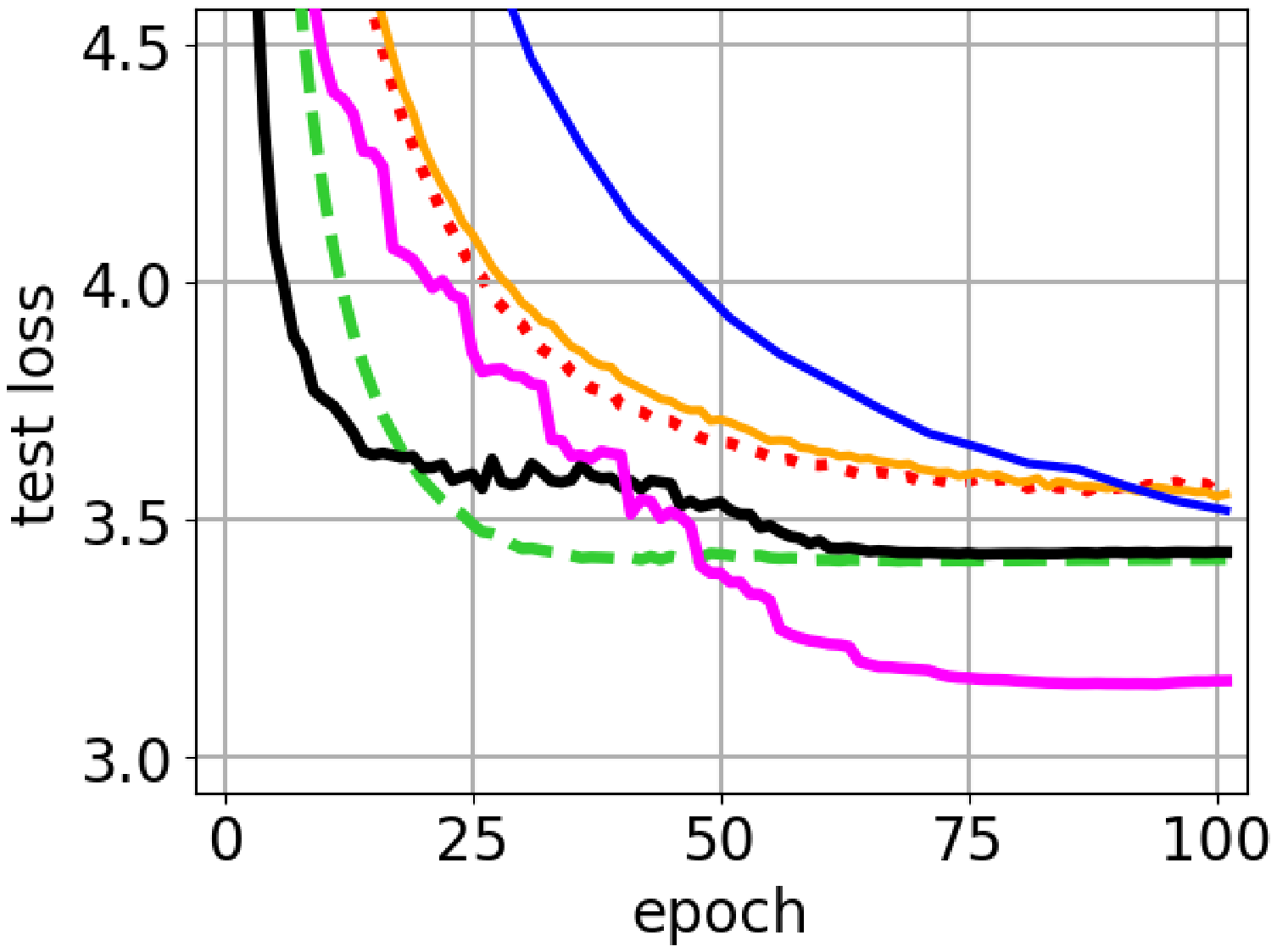} &
\includegraphics[width=\fw]{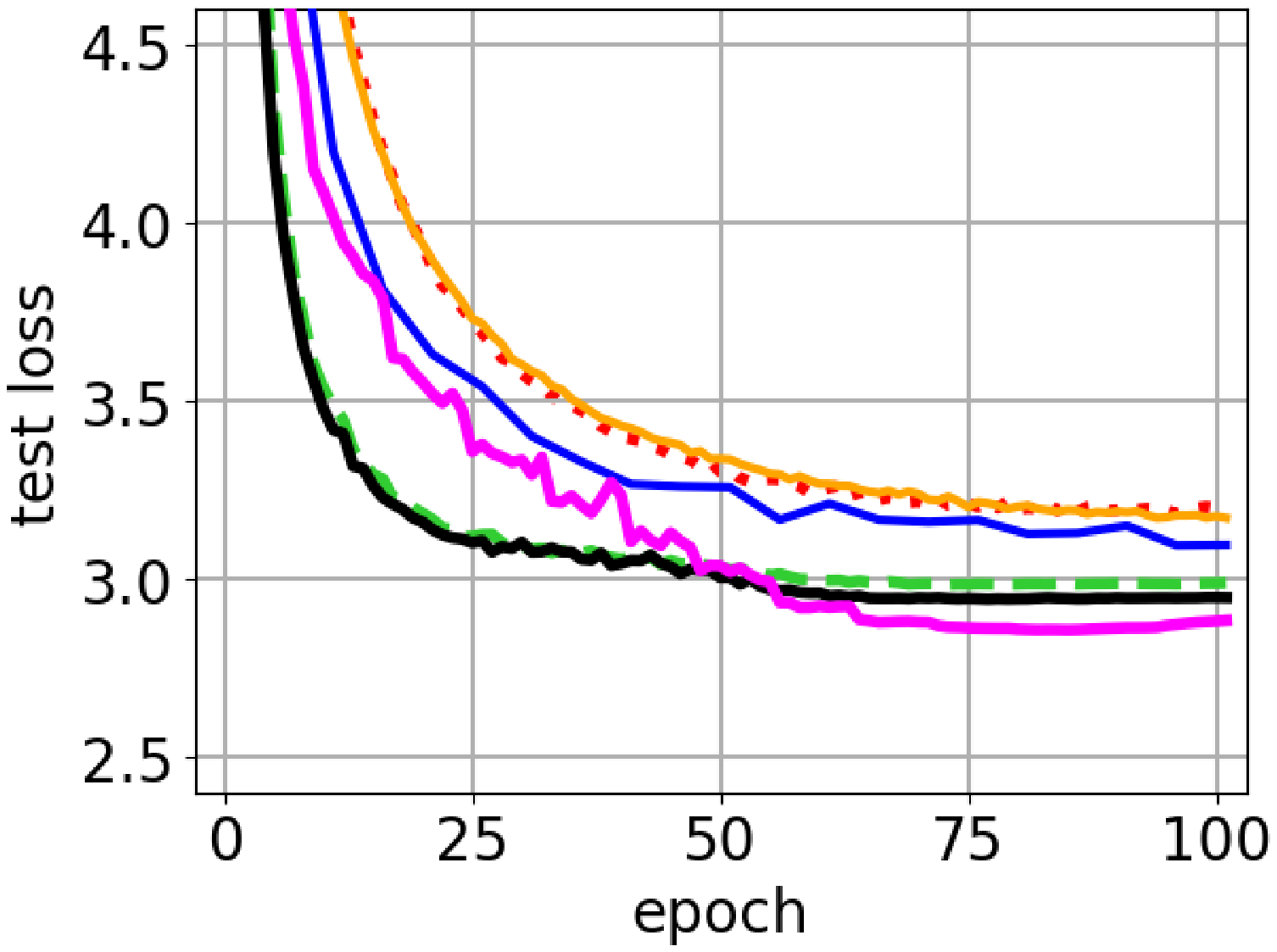} &
\includegraphics[width=\fw]{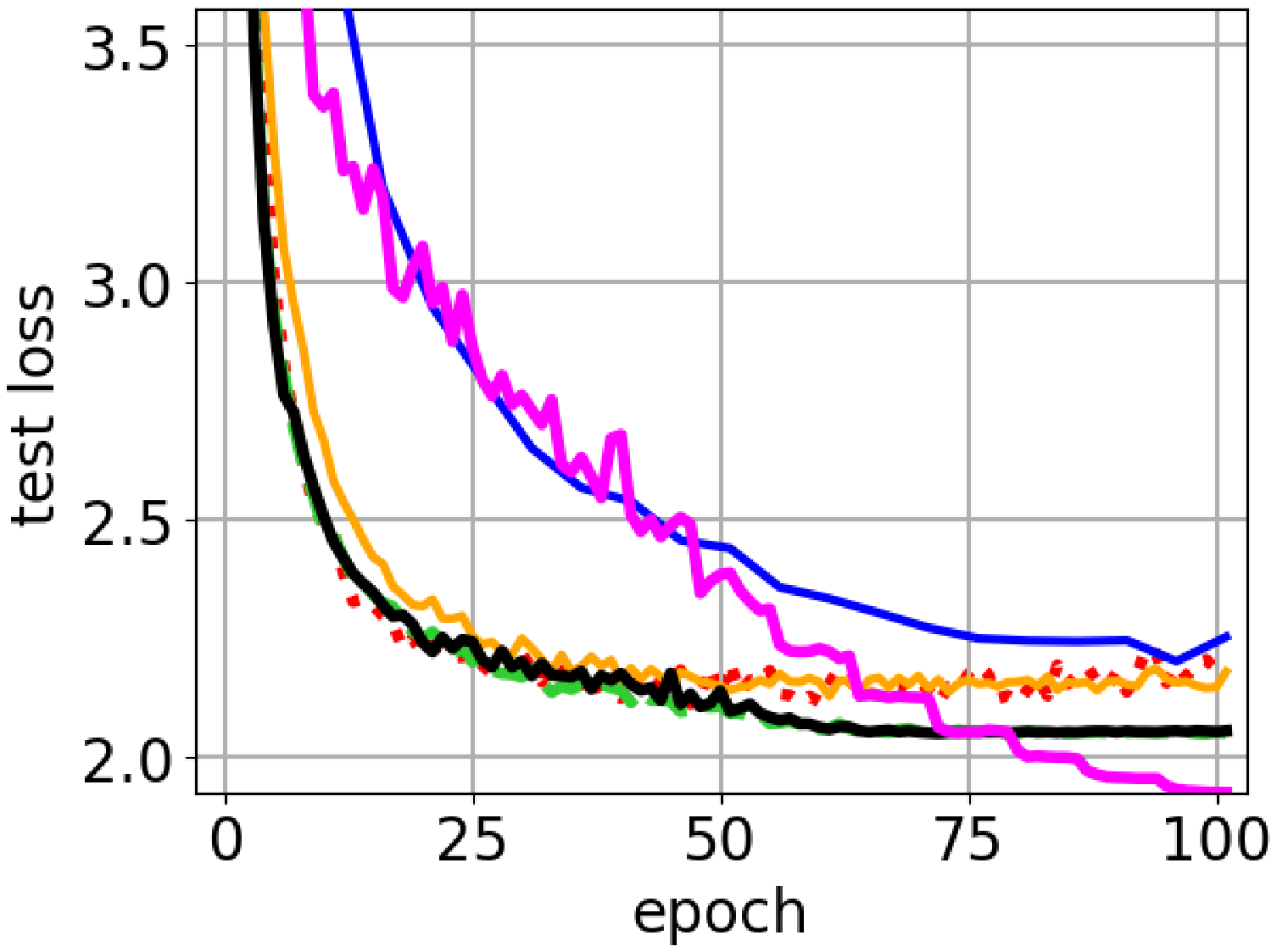} \\
& \multicolumn{1}{c}{\qquad NN-2} & \multicolumn{1}{c}{\qquad NN-3} & \multicolumn{1}{c}{\qquad LeNet} \\
\end{tabular}
\vspace{-5pt}
\caption{
Test loss curve for MNIST (top), Fashion-MNIST (middle), and EMNIST-Letters (bottom) over epoch (x-axis) by (1st, 2nd, 3rd columns) NN-2, NN-3, LeNet optimized using SGD (black line), RMSprop (red), Adam (yellow), Entropy-SGD (blue), Accelerated-SGD (green), and ours (magenta) using the linear trimming with $\tr=20\%$: The test loss is averaged within individual 10 trials.}
\label{fig:coomparion_to_others:curves}
\end{figure}
%
%
%
%
\par
Furthermore, we extend our experiment to NN-2, NN-3, and LeNet with MNIST, Fashion-MNIST, and EMNIST-Letters data-sets.
Figure~\ref{fig:experiment:ours:loss} presents the test loss by the network models optimized using
a range of the label noise of $\nr=0, 2.5\%, 5\%, 7.5\%, 10\%, 15\%$ 
in combination with the example trimming with $\tr=0$ (SGD), and ours with $\tr=10\%,20\%,40\%$.
%
As shown in Figure~\ref{fig:experiment:ours:loss}, 
the trimming rate $\tr$ in proportional to the noise ratio $\nr$ achieves favorable curve of the test loss.
%
Figure~\ref{fig:experiment:ours:loss} also demonstrates ours is robust to the choice of the trimming ratio $\tr$.
We use $\tr=20\%$ as the recommended condition in the following experiment.
\subsection{Comparison to the state-of-the-arts}
We now compare our algorithm with the state-of-the-art optimization methods: 
SGD, 
RMSprop~\cite{tieleman2012lecture} (RMS),
Adam~\cite{kingma2014adam},
Entropy-SGD~\cite{Chaudhari2017EntropySGD} (eSGD),
Accelerated-SGD~\cite{Kidambi2018Acc} (aSGD).
We impose the label noise with $\nr=0, 2.5\%, 5\%, 7.5\%, 10\%, 15\%$ to the tested algorithms, and choose their best results for a fair comparison.
Regarding hyper parameters, 
our algorithm uses the trimming ratio of Eq.(\ref{eq:linear-trimming}) with $\tr=20\%$.
We use grid search and set 0.95 as the weighting factor in RMSprop, 0.9 and 0.999 for the first and second momentum factors in Adam.
The hyper-parameters for eSGD and aSGD are also set as the recommended in the original papers including the Langevin loop number of 5 for eSGD.
The sigmoid learning-rate annealing is applied to non-adaptive methods: SGD, eSGD, aSGD and ours.
We choose one of the four learning-rate scales for each condition such that each algorithm has achieved the best result in the last $10\%$-epoch test loss.
\par
Figure~\ref{fig:coomparion_to_others:curves} presents the test loss curve over epoch of the optimization methods.
Our algorithm has drawn better test-loss curves in the last half of epochs.
This demonstrates that our algorithm has successfully escaped local points where the other algorithms have been trapped.
Table~\ref{tab:comparion_to_others:loss} summarizes the test accuracy, where we present both the mean and minimum test loss by the optimization methods.
Note that SGD employs the sophisticated learning-rate annealing with sigmoid function and has achieved better or comparable results with the other state-of-the-art optimizations.
%
As shown in Table~\ref{tab:comparion_to_others:loss}, the proposed algorithm has achieved better or comparable test loss irrespective of the data-sets with different number of classes and the model architectures.
%
%
%
%
%
%
%
%
%
\section{Conclusion} \label{sec:conclusion}
We have proposed a first-order optimization method that uses the label noise with the example trimming.
The example trimming in our algorithm avoids outliers degenerating the model and enables us to use a relatively large label-noise.
As the result, our algorithm imposes an implicit regularization that improves generalization of the trained network.
Due to the nature of the label noise, our algorithm is independent to both the model structure and data-sets.
The effectiveness of the proposed algorithm has been demonstrated by the experimental results in which our algorithm overtakes a number of the state-of-art optimization methods in the image classification task.
\par
A limitation of our work is as follows.  
In order to experiment more deep networks, 
we have also tested our method based on CIFAR-10~\cite{krizhevsky2009learning} that is one of the major benchmark for the image classification using color images.  
However, the test loss by our method was inferior to the baseline SGD even applied to NN-2 networks.
This implies that if the dimension of input is larger than those we have tested, an additional inspection is necessary to make further progress, yet the proposed method would be directly beneficial for application studies using the one-dimensional data, e.g., medical images and depth data combined with the neural networks.
\section*{Acknowledgements}
This work was supported by NRF-2019K1A3A1A77074958, the Korea government (MSIT): IITP-2021-0-01341, Artificial Intelligence Graduate School, Chung-Ang University and IITP-2021-0-01574, High-Potential Individuals Global Training Program.
\bibliography{image201216}
\end{document}